\documentclass[10pt,twocolumn,letterpaper]{article}

\usepackage[pagenumbers]{cvpr} %

\usepackage[dvipsnames]{xcolor}

\definecolor{cvprblue}{rgb}{0.21,0.49,0.74}
\usepackage[pagebackref,breaklinks,colorlinks,citecolor=cvprblue]{hyperref}

\usepackage{graphicx}
\usepackage{multicol}
\usepackage{amsmath}
\usepackage{amssymb}
\usepackage{booktabs}
\usepackage{listings}
\usepackage{booktabs, multirow} %
\usepackage{soul}%
\usepackage{changepage,threeparttable} %

\usepackage[pagebackref,breaklinks,colorlinks]{hyperref}

\usepackage[capitalize]{cleveref}

\usepackage{amsmath,amsfonts,bm}

\def\eqref#1{equation~\ref{#1}}

\def\1{\bm{1}}

\DeclareMathAlphabet{\mathsfit}{\encodingdefault}{\sfdefault}{m}{sl}
\SetMathAlphabet{\mathsfit}{bold}{\encodingdefault}{\sfdefault}{bx}{n}

\usepackage{svg}

\usepackage{adjustbox}
\usepackage{algorithm}
\usepackage{algorithmic}
\usepackage{amsfonts}
\usepackage{amsmath}
\usepackage{array}
\usepackage{booktabs}
\usepackage{caption}
\usepackage{color}
\usepackage{enumitem}
\usepackage{fancyvrb}
\usepackage{float}
\usepackage[symbol]{footmisc}
\usepackage{inconsolata} %
\usepackage{listings}
\usepackage{longtable}
\usepackage{makecell}
\usepackage{mathtools}
\usepackage{multirow}
\usepackage{subcaption}
\usepackage{supertabular}
\usepackage{tabto}
\usepackage{tablefootnote}
\usepackage{tabularx}
\usepackage{verbatim}
\usepackage{xfrac}
\usepackage{xltabular}

\renewcommand{\thefootnote}{\fnsymbol{footnote}}

\newenvironment{proof-of}[1]{{\em Proof of #1:}}{}

\everypar{\looseness=-1}

\newcommand{\iiitdlogo}{\raisebox{5pt}{\includegraphics[scale=0.0113]{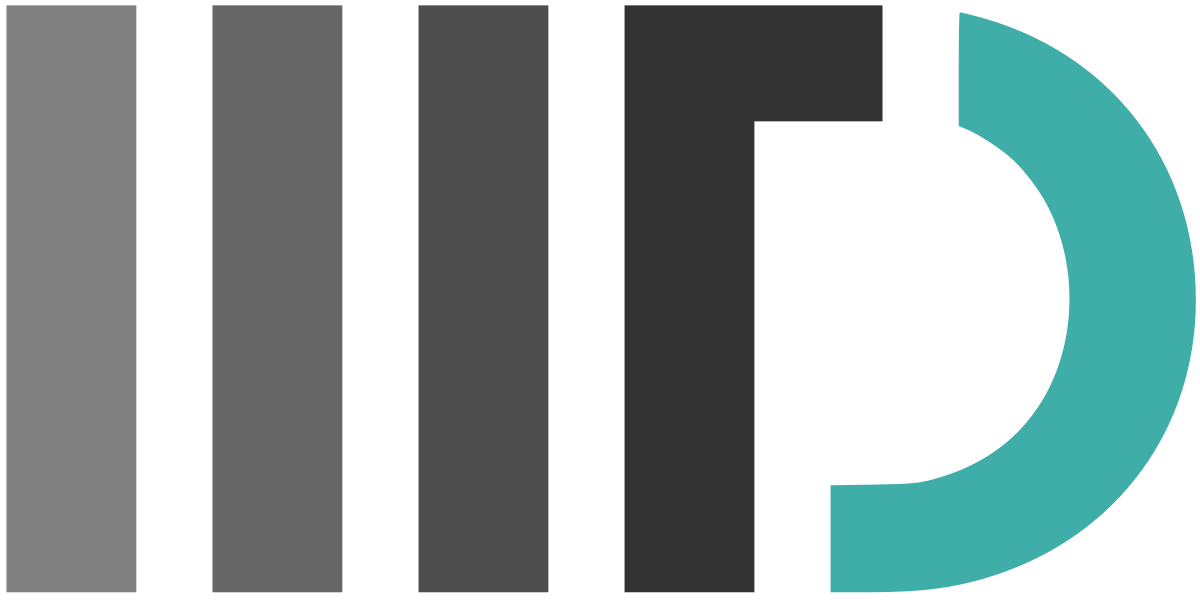}}}
\newcommand{\ublogo}{\raisebox{5.2pt}
{\includegraphics[scale=0.085]{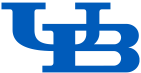}}}

\newcommand{\adobelogo}{\raisebox{5pt}{\includegraphics[scale=0.032]{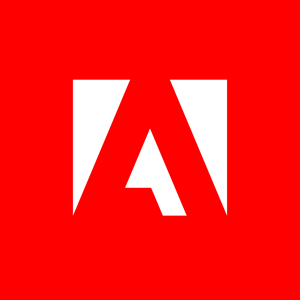}}}
\newcommand{\mbzuailogo}{\raisebox{5pt}{\includegraphics[scale=0.062]{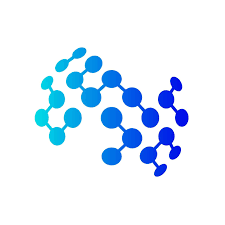}}}
\newcommand\coauth{$^\star$}
\newcommand\blfootnote[1]{%
  \begingroup
  \renewcommand\thefootnote{}\footnote{#1}%
  \addtocounter{footnote}{-1}%
  \endgroup
}

\usepackage[skins]{tcolorbox} %

\definecolor{valbest}{HTML}{d9ead3}
\newcommand{\valbest}[1]{\colorbox{valbest}{#1}}
\definecolor{valgood}{HTML}{d0e0e3}
\newcommand{\valgood}[1]{\colorbox{valgood}{#1}}
\definecolor{valmid}{HTML}{fce5cd}

\definecolor{valbad}{HTML}{ead1dc}

\definecolor{themegreen}{HTML}{365956}
\definecolor{themepurple}{HTML}{3c1b48}
\definecolor{themered}{HTML}{b43748}

\newcolumntype{L}[1]{>{\raggedright\let\newline\\\arraybackslash\hspace{0pt}}m{#1}}
\newcolumntype{C}[1]{>{\centering\let\newline\\\arraybackslash\hspace{0pt}}m{#1}}
\newcolumntype{R}[1]{>{\raggedleft\let\newline\\\arraybackslash\hspace{0pt}}m{#1}}

\setlist[itemize]{noitemsep, topsep=0pt}
\setlist[enumerate]{noitemsep, topsep=0pt}

\crefname{section}{Sec.}{Secs.}
\Crefname{section}{Section}{Sections}
\Crefname{table}{Table}{Tables}
\crefname{table}{Tab.}{Tabs.}

\newcommand{\cy}[1]{}%

\def\paperID{2648} %
\def\confName{CVPR}
\def\confYear{2024}

\begin{document}

\title{\vspace*{-10mm}Behavior Optimized Image Generation\vspace*{-5mm}}

\author{Varun Khurana \coauth \adobelogo, Yaman K Singla \coauth\adobelogo  \hspace{0.1mm} \ublogo \hspace{0.1mm} \iiitdlogo, Jayakumar Subramanian\adobelogo,\\ Rajiv Ratn Shah\iiitdlogo, Changyou Chen\ublogo, Zhiqiang Xu\mbzuailogo, Balaji Krishnamurthy\adobelogo\\
\adobelogo Adobe Media and Data Science Research, \ublogo SUNY at Buffalo, \iiitdlogo IIIT-Delhi, \mbzuailogo MBZUAI\\
}
\maketitle
\blfootnote{\coauth \small Equal Contribution. Contact varunkhurana@adobe.com, ykumar@adobe.com for questions and suggestions.}
\begin{abstract}
   The last few years have witnessed great success on image generation, which has crossed the acceptance thresholds of aesthetics, making it directly applicable to personal and commercial applications. However, images, especially in marketing and advertising applications, are often created as a means to an end as opposed to just aesthetic concerns. The goal can be increasing sales, getting more clicks, likes, or image sales (in the case of stock businesses). Therefore, the generated images need to perform well on these {\em key performance indicators} (KPIs), in addition to being aesthetically good. In this paper, we make the first endeavor to answer the question of ``How can one infuse the knowledge of the end-goal within the image generation process itself to create not just better-looking images but also ``better-performing'' images?''. We propose BoigLLM, an LLM that understands both image content and user behavior. 
   BoigLLM knows how an image should look to get a certain required KPI. We show that BoigLLM outperforms 13x larger models such as GPT-3.5 and GPT-4 in this task, demonstrating that while these state-of-the-art models can understand images, they lack information on how these images perform in the real world. To generate actual pixels of behavior-conditioned images, we train a diffusion-based model (BoigSD) to align with a proposed BoigLLM-defined reward. We show the performance of the overall pipeline on two datasets covering two different behaviors: a stock dataset with the number of forward actions as the KPI and a dataset containing tweets with the total likes as the KPI, denoted as {\em BoigBench}. To advance research in the direction of utility-driven image generation and understanding, we release BoigBench, a benchmark dataset containing 168 million enterprise tweets with their media, brand account names, time of post, and total likes.
\end{abstract}

\begin{figure*}
  \centering
  \includegraphics[width=1.0\textwidth]{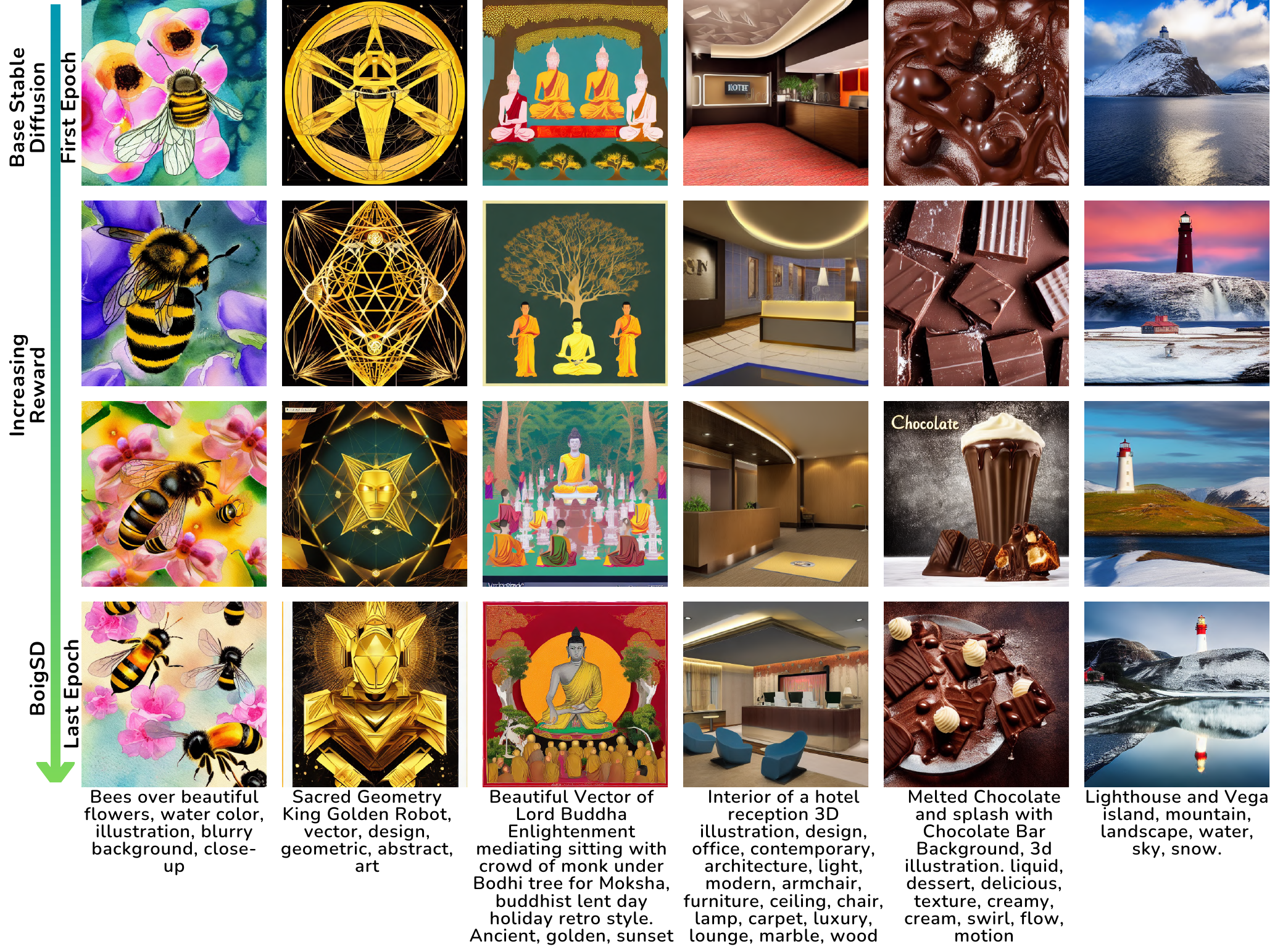}
  \caption{Progression of BoigSD generated images over the course of training: Top-1 images selected by BoigLLM-based reward out of 64 generations. BoigLLM optimizes for image utility (user behavior) over the course of training. While aesthetics and other metrics like prompt adherence may change over generations, the key change that BoigLLM brings is the optimization of the end-objective, \textit{i.e.}, an improvement in the downstream image KPI. The end objective, in our case, is the probability of achieving a higher probability on the objective of KPI (=High). Here, we see that BoigLLM also improves aesthetics and prompt adherence for a few prompts. See Fig.~\ref{img:comparison-marketing-image} for more marketing samples generated by BoigSD based on prompts from enterprise tweets (sourced from BoigBench). \label{fig:journey of boigSD images}}
\end{figure*}

\section{Introduction}
\label{sec:intro}
A number of ML-based applications are built to optimize human behavior as the end goal; for example, recommendation systems are optimized to get maximum \textit{interest} \cite{bobadilla2013recommender}, search engine retrieval is optimized to get to relevant documents to a user's query \textit{faster} \cite{menick2022teaching}, game-play is optimized for increasing complexity to keep humans \textit{engaged} \cite{nadiger2019federated}, robotic control is optimized for copying human locomotion \cite{christiano2017deep}, chat-engines are optimized for being \textit{helpful} \cite{ouyang2022training}, and summarization systems are optimized to keep those sentences in the summary which humans consider as important \cite{stiennon2020learning}. Today images are generated to be more real-looking \cite{dhariwal2021diffusion,saharia2022photorealistic,ho2020denoising,rombach2022high} and aesthetical \cite{xu2023imagereward,kirstain2023pick,black2023training}. Advances in fulfilling this objective have brought image generation technology to a threshold such that it is being deployed as a widely used commercial application. The next step in the evolution is to optimize the image generation with the goal of the image-maker. The utility of an image is better communication. The images are, therefore, a means to an end. For instance, in the case of marketing, an image's utility is to bring in more customer engagement measured through metrics like clicks, likes, shares, and comments, ultimately leading to more sales and brand building. In this paper, we make initial efforts to answer the question of how one can infuse the information of utility within the image generation process itself to create not just better-looking images but also better-performing images.

\begin{figure}[h]
  \centering
  \includegraphics[width=0.5\textwidth]{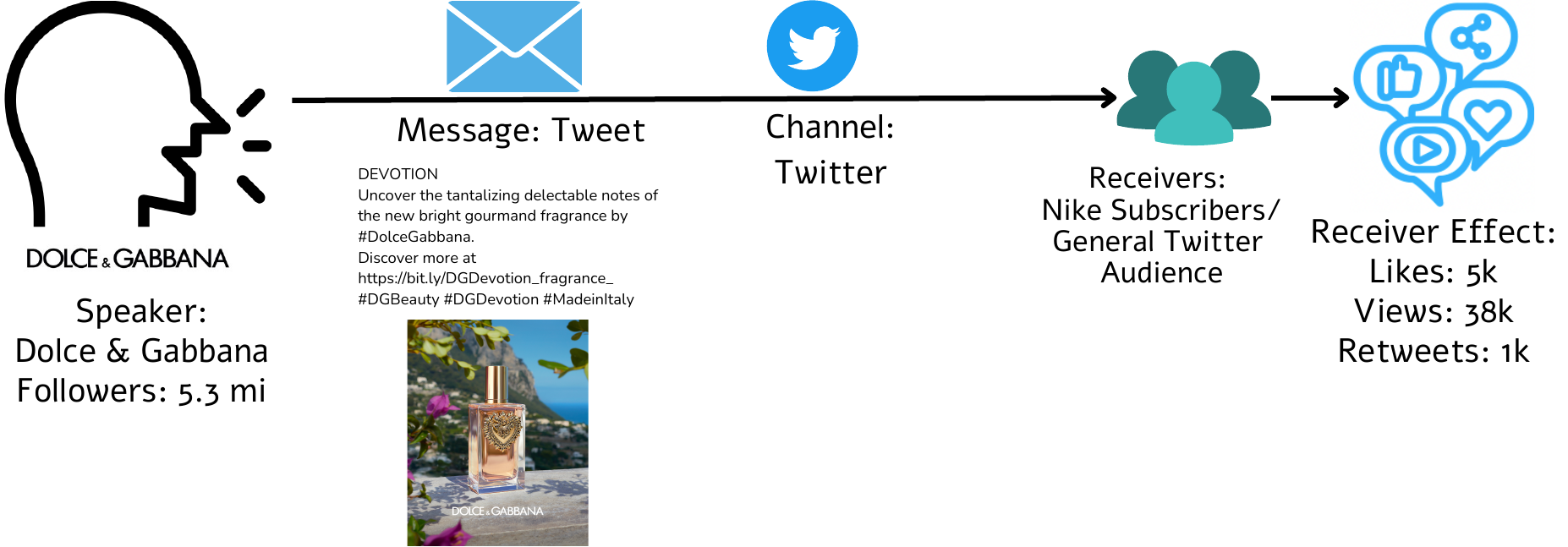}
  \caption{Any message is created to serve an end goal. For marketers, the end goal is to bring in the desired receiver effect (behavior) (like clicks, purchases, likes, and customer retention). The figure presents the key elements in the communication pipeline - the marketer, message, channel, receivers, and finally, the receiver effect. Traditionally, image generation is optimized on metrics such as aesthetics and FID. For effective communication, the image generation process needs to be optimized on the receiver effect (other than the traditional metrics).  \label{fig:factors-of-communication}}
\end{figure}

Images, in the form of messages like text, can be used as a means of communication. A receiver, upon receiving a (image-based) message from a sender, interacts with the message, thereby generating effects (analytics) such as likes, comments, shares, and purchases (Fig.~\ref{fig:factors-of-communication}). Just about every transaction involving an image can be formulated as a transaction between a sender and a receiver over a channel generating some effect (behavior) \cite{shannon-weaver-1949,lasswell1948structure}. The eventual goal of a sender is to get the desired effect \cite{shannon-weaver-1949,khandelwal2023large}. Therefore, it makes sense to use the \textit{effect} as the optimization goal of the image generation process. Today, in the absence of such a technology that can directly generate performant images, savvy marketers perform the task of generating high-performing images as a two-step process. In the first step, marketers ask artists to make several competing images with certain characteristics and then perform A/B tests over them to find out which is working for their customer base. A/B tests are expensive in terms of time, money, and messaging dissonance costs between their customer base and can be conducted only in small numbers due to their operational nature \cite{bhat2020near}. These costs force marketers to avoid this tedious process altogether and instead go with their heuristics (feel) instead of scientifically backed optimization. In this paper, we propose to merge the two-step process into a single step and induce the goal of generating performant images within the diffusion process itself. 

\begin{figure*}[h]
  \centering
  \includegraphics[width=1.0\textwidth]{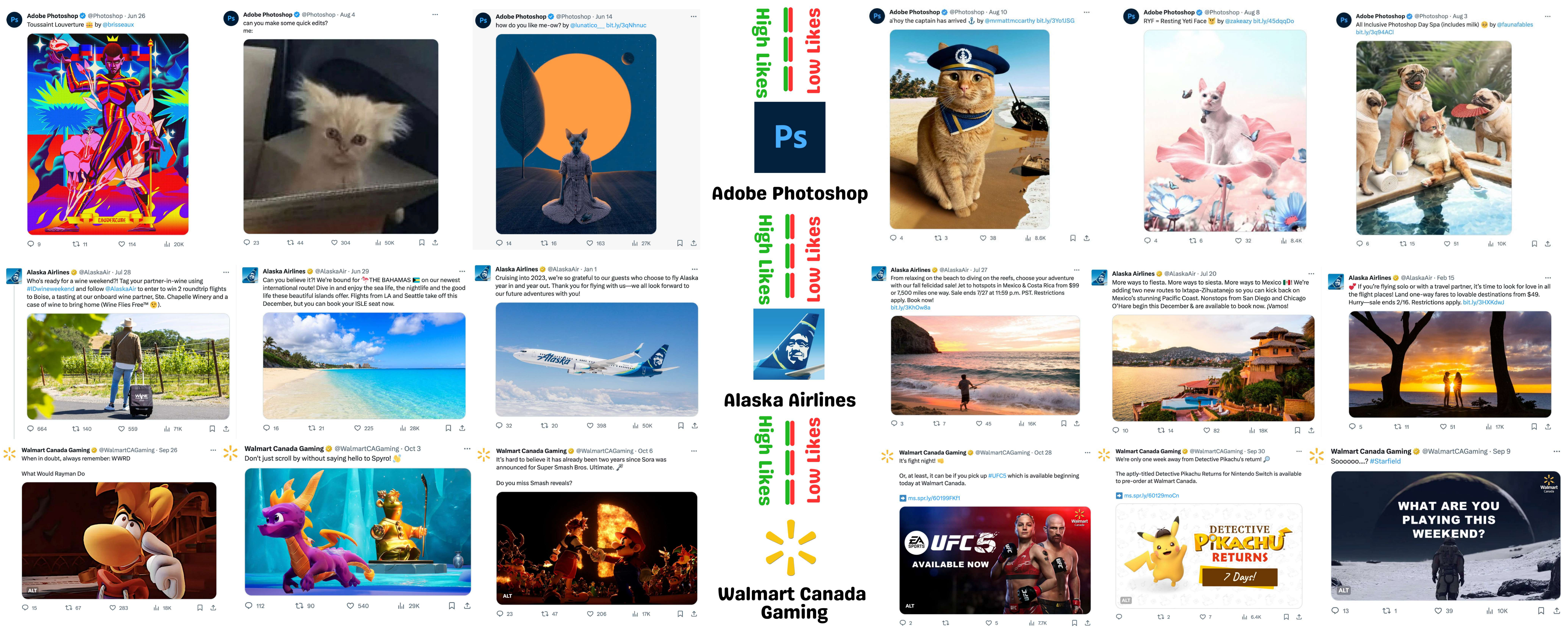}
  \caption{Sample media and tweets from enterprise accounts in the BoigBench dataset. It can be noted, for example, in the Adobe Photoshop tweets, that the media does not differ significantly in aesthetics or objects themselves (all of them are cats). Despite that, there is much difference in the image KPIs, indicating that utility as desired user behavior is separate from other optimization objectives such as aesthetics or prompt adherence.\label{fig:boigbench-dataset-samples}}
\end{figure*}

To design a model that can generate natural language descriptions of images based on their potential to elicit different effects, we design a large language model (LLM) that understands both images (content) and receiver behavior, such as likes and downloads. We call this model BoigLLM. To train BoigLLM, we present four types of behavior fine-tuning \cite{khandelwal2023large} (\S\ref{sec:boigLLM}, Listing~\ref{BoigLLM:verbalization-1}-\ref{BoigLLM:verbalization-4}) to train the model to predict image attributes for a given communicator, time, channel, topic, and intended behavior. We test its performance with respect to state-of-the-art language models (GPT-3.5 \cite{ouyang2022training}, GPT-4 \cite{openai2023gpt4}) and fine-tuned Llama \cite{touvron2023llama2}, and find that BoigLLM outperforms all other models despite being at least 13x smaller in size than other models. Curiously, while in-context learning over GPT-3.5 and 4 has been shown to work well in other domains \cite{min-etal-2022-rethinking}, it does not work in behavior-conditioned generation (Table~\ref{tab:twitter_boigbench_data_boig_llm_results}). Once trained, BoigLLM can generate natural language descriptions of how an image conditioned on behavior should look like. Further, BoigLLM's logits act as the simulator to test the effectiveness of an image in real-world settings (\S\ref{sec:Performance Aligning Stable Diffusion}).

Next, in order to generate actual pixels of behavior-optimized image generation, we take motivation from literature about supervised fine-tuning (SFT) and recent reinforcement learning with human feedback (RLHF) in computer vision and natural language processing. SFT and RLHF present two competing techniques, which are effective in two different feedback regimes - feedback involving the correct response and scoring the generated response, respectively, to induce desired characteristics in the output. SFT in diffusion models has been used to bias diffusion for a certain subject \cite{ruiz2023dreambooth}, a certain style \cite{lu2023specialist} or conditional inputs \cite{zhang2023adding}. RLHF, a more recent method, has been used to optimize for objectives such as higher aesthetics \cite{black2023training,clark2023directly}, higher compressibility \cite{black2023training,clark2023directly}, to bias diffusion towards certain object classes or colors \cite{lee2023aligning,clark2023directly} or implicit human preferences \cite{kirstain2023pick,xu2023imagereward}.

Following the two techniques, we design experiments to show the impact of both on the downstream task of behavior-optimized image generation. To make the model learn about behavior using SFT, we finetune it on high effect-generating samples. We find that a simple supervised fine-tuning approach is unable to increase the intended behavior of images (Table~\ref{tab:twitter_data_scores}). For RLHF, we use rewards obtained using BoigLLM and train the stable diffusion model using denoising diffusion policy optimization (DDPO) \cite{black2023training}. DDPO uses the fact that the denoising process in diffusion models is a Markov process and hence standard RL approaches such as a policy gradient based approach can be used to modify it to yield some desired objective such as aesthetic quality. This objective needs to be specified in the form of a reward function to be used in DDPO, which in our case, is the BoigLLM-based reward model.
We find that stable diffusion aligned with rewards given by BoigLLM performs better than both fine-tuned and base stable diffusion models.
We test the overall pipeline on two datasets capturing two different behaviors (\S\ref{ref:setup}): an image stock dataset with the number of downloads as the key performance indicator (KPI) and a dataset consisting of tweets with total likes as the KPI (BoigBench).

Further, while recent advances have been made both in language modeling \cite{kreutzer2018can,stiennon2020learning,ziegler2019fine,nakano2021webgpt} and computer vision to align models with user expectations, there is a huge dearth of large and open datasets which can help align generation models better. Especially in the case of text-to-image models, there have been only a few recent works that release human preferences related to count, color, background \cite{lee2023aligning}, text coherence \cite{xu2023imagereward}, and aesthetics \cite{pressman2023simulacra,wu2023better,kirstain2023pick}. These pioneering works, while providing direction and data to try out human alignment for text-to-image generation models, capture the operational \cite{lee2023aligning,xu2023imagereward} and artistic \cite{pressman2023simulacra,wu2023better,kirstain2023pick} elements of image generation as a process, as opposed to the utility of images generated. Therefore, to spur research in the direction of utility-drive image generation, we release a large-scale dataset, BoigBench. BoigBench consists of 168 million tweets capturing 17 years of enterprise content for over ten thousand brand accounts and content's associated utility in terms of average user behavior of tweet likes\footnote{The dataset was collected using Twitter academic API over a period of several years.}. BoigBench serves as a starting point for modeling large scale behavior aligned image generation. We cover more details of the dataset in Sec.~\ref{sec:BoigBench}.

\noindent To summarize, we make the following contributions:

1. We introduce the problem of behavior-optimized image generation (BOIG) and present several approaches, making some initial strides in solving this problem. Images, especially in advertising settings, are created as a means to an end. Therefore, the image generation process needs to be biased on the image's eventual utility (along with being of high aesthetics and fidelity). First, we demonstrate that state-of-the-art language models (GPT-3.5, GPT-4, Llama) and the stable diffusion model, while having shown the ability to understand and generate images, lack information about how they would perform in the real world. Further, while in-context learning has been shown to work in many NLP domains, does not work in behavior-conditioned image description generation (Table~\ref{tab:twitter_boigbench_data_boig_llm_results}). 

2. We introduce behavior fine-tuning approach to fine-tune an LLM to teach it behavior and content together. We call it BoigLLM. BoigLLM, once trained, can generate image description conditioned on the needed behavior, communicator, and time. We show that BoigLLM outperforms GPT-3.5, GPT-4, and fine-tuned Llama on behavior conditioned image description generation (Table~\ref{tab:twitter_boigbench_data_boig_llm_results}).

3. We finetune stable diffusion with BoigLLM-provided rewards to create BoigSD, which can generate progressively higher utility images. BoigSD outperforms base stable diffusion on two datasets (image stock and BoigBench) (Table~\ref{tab:twitter_data_scores}) (Figs.~\ref{fig:journey of boigSD images},\ref{img:comparison-marketing-image}). Further, we show that simply training diffusion model with high-KPI images does not work and, in fact, performs worse than base stable diffusion. This provides insights into how behavior optimization as the utility of image generation can be integrated into generative models.

4. \textbf{BoigBench:} To further the goal of behavior-optimized image generation, we release the first large-scale dataset, BoigBench, consisting of images and corresponding human behavior. BoigBench consists of 168 million tweets collected from 10135 enterprise Twitter accounts from the time period 2007 to 2023. Each row consists of the account name, tweet text, media posted with the tweet, image captions, keywords, colors and tones, the time of the tweet, and the number of likes the tweet received. This is the first dataset to enable behavior-optimized image generation and adds to the efforts of the few other datasets used for aligning diffusion models with human preferences \cite{kirstain2023pick,xu2023imagereward}.

\section{Setup}
\label{ref:setup}
In this section, we cover the details of the problem setup, talking about the pipeline for the systematic collection of images and their performance data. %

\subsection{BoigBench}
\label{sec:BoigBench}
To gain insights into image performance and align text-to-image generation with performance, we start by collecting a large dataset of image performance.
Our data collection method involved leveraging Twitter, a platform extensively utilized by brands for various purposes such as ongoing product campaigns, sales, offers, discounts, brand building, and community engagement \cite{alalwan2017social}. Twitter user engagement metrics encompass user likes, retweets, comments, mentions, follows, clicks on embedded media, and links. 
However, the Twitter API provides access to only user likes, retweets, and comments for a given post, with access to comments necessitating a separate and costly call. Therefore, utilizing academic API licenses, we extracted the following data from Twitter: Tweet ID, company name, username, timestamp, tweet text, media files, and user likes.

We began by compiling a comprehensive list of company names using the Wikidata knowledge graph \cite{wikidata}, focusing on entities categorized as `business' or `enterprise.' Subsequently, we conducted Google searches to gather a list of all associated accounts for these companies. For example, for Adobe, this encompassed accounts like Adobe, Adobe Photoshop, Adobe Lightroom, Adobe Experience Cloud, and so forth. This method enabled us to amass a total of 10,135 Twitter handles. We then utilized the Twitter API to retrieve tweets posted by these enterprises spanning from 2007 to the closure of the Twitter API in January 2023. This effort resulted in the collection of 168 million tweets over a 17-year period, with 28.5 million of these tweets featuring various forms of media, including GIFs, images, and videos. Fig.~\ref{fig:boigbench-dataset-samples} shows several examples of media and tweets present in the BoigBench. Among tweets containing media, the average number of media items per tweet was 2.04. Figures~\ref{fig:company_tweets_vs_time} depict the distribution of likes and the number of posts over time, revealing the absence of significant seasonal patterns in the distribution of likes. Further, Figs.~\ref{fig:aesthetic_score_distribution_stock},\ref{fig:aesthetic_score_distribution_boigbench} shows the distribution of aesthetics between high and low KPI images, demonstrating that there is no significant difference in aesthetics between high and low KPI images.

We sample 79,728 tweets posted in the last five years by 982 Twitter handles. For this, the top 10 and the bottom 60 percentile of all tweets per account were sampled based on the number of user likes, such that both buckets have roughly an equal number of media samples. These buckets are subsequently referred to as `high' and `low' KPI buckets, respectively. Since it is hard to assign KPI credit to the multiple media present in a single tweet, we assign an equal KPI credit to all the media in a tweet. This resulted in 35,649 high-KPI images and 44,079 low-KPI images. The high-KPI tweets had an average of 5253.04 likes, while low-KPI tweets had an average of 3.06 likes. Subsequently, all images were verbalized by extracting their captions using LlaVA \cite{liu2023visual}, colors and tones along with their coverage using \cite{Qin_2020_PR}, and object tags using RAM \cite{zhang2023recognize} which are further fed to Grounding DINO \cite{liu2023grounding} to obtain the object bounding boxes. We randomly sampled 1000 samples from each bucket for testing, with the remaining samples used for training. Table~\ref{tab:twitter_data_scores-ground-truth} presents an overview of the ground truth data in BoigBench in terms of aesthetic score, CLIP score, and number of objects.

\subsection{Stock Dataset}
\label{sec:Stock Dataset}

In addition to the BoigBench dataset, we incorporated another dataset to evaluate the performance of utility-driven image generation. This dataset is sourced from a Stock business selling images. The dataset contains images, their captions and keywords, and their associated KPIs as the number of downloads, ``forward actions'', and impressions. Forward action is a type of non-purchase engagement metric recorded after a user clicks, paginates, or raises a license request after searching for an asset. Naturally, being non-purchase metrics, the number of forward actions and impressions is much more than the number of downloads. To create balanced subsets, we leveraged the relatively less skewed forward actions metric to divide the data into three approximately equal-sized buckets.

In the end, the high bucket contained 14,649 samples; the medium bucket had 12,834, and 20,142 in the low bucket. These samples include captions, keywords, release dates, image content, and the three KPIs. To create a test set, we randomly selected 1000 samples from each bucket, while the remaining samples were designated for the training set.

\section{Method}
\label{sec:method}
We adopt three methods to fulfill the goal of behavior-optimized image generation, the first one working in the natural language domain, generating a description of how a behavior-optimized image should look like, and the other two working in the vision domain, generating actual behavior-optimized pixels. 
We cover each of them next.

\subsection{BoigLLM}
\label{sec:boigLLM}
Prior works \cite{bhattacharya2023video} have demonstrated the capability of language-only pre-trained models like GPT-3 and Vicuna to infer information about visual content without explicit visual reasoning training. Recent models such as BLIP \cite{li2023blip2}, Llava \cite{liu2023visual}, MiniGPT-4 \cite{zhu2023minigpt}, and GPT-4 \cite{openai2023gpt4} have shown language models' ability to 'see' by incorporating visual branches (often a combination of ViT \cite{dosovitskiy2020image} and Qformer \cite{li2023blip2}) and training them with image-language instructions to answer image-related questions. However, our findings (Table~\ref{tab:twitter_boigbench_data_boig_llm_results}) reveal that neither pertaining nor further visual instruction tuning gives a language model the ability to simulate the downstream behavior of an image-based communication or reason about how a more performant image should look like. Further, we also find that in-context learning, while successful in many other domains, does not perform well in behavior-related tasks. Therefore, to teach a language model about image content and downstream performance, we further train the Llama LLM.

\begin{figure*}[h]
  \centering
  \includegraphics[width=0.95\textwidth]{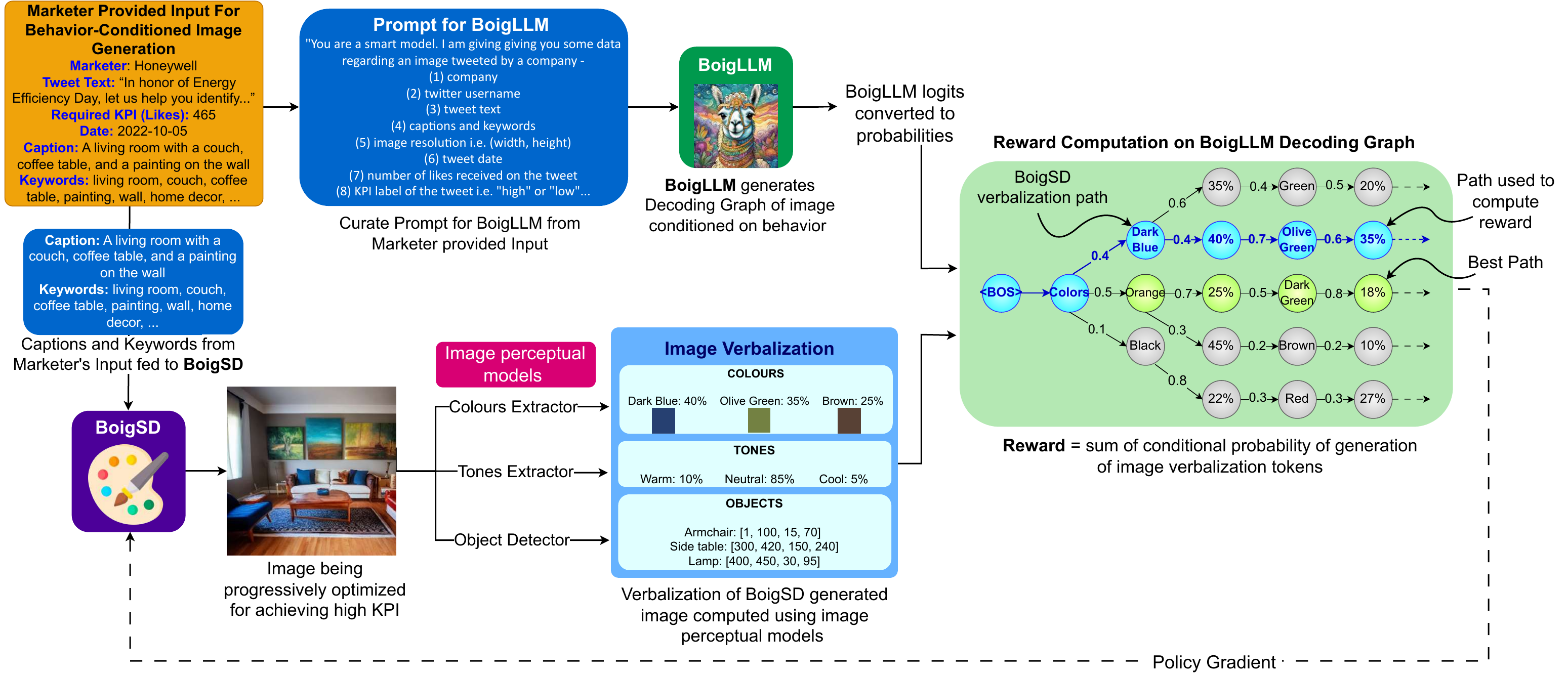}
  \caption{\label{fig:architecture_diagram}
  Architecture of the proposed pipeline for behaviour-optimised image generation. Once trained, BoigLLM possesses the capability to generate verbal descriptions of an image based on conditioning factors such as the marketer, time, caption, keywords, tweet text, and the specified Key Performance Indicator (KPI). BoigLLM inherently understands how an image should align with a given KPI and prompt.
BoigSD takes a prompt and generates an image, which then undergoes verbalization via image perception models. Its objective is to create images that, when verbalized, closely resemble the behavior-conditioned verbalization generated by BoigLLM. The verbalized output of BoigSD is fed into the reward model. Using the logits provided by BoigLLM, a reward is computed for BoigSD, indicating how closely this verbalized output aligns with BoigLLM. This reward value serves as feedback for BoigSD in the form of policy gradient, aiding in its continual improvement and refinement within the image generation process. Thus, this pipeline trains BoigSD to generate behavior-optimized images by gradually aligning its output with BoigLLM. \vspace*{-3mm}
  }
\end{figure*}

To teach Llama about an image and its downstream KPI, we perform behavior fine-tuning \cite{khandelwal2023large}. We design four types of behavior-finetuning instructions (Listings~\ref{BoigLLM:verbalization-1}-\ref{BoigLLM:verbalization-4}). The idea is to \textit{verbalize} an image using image perception models like color extractor, tones extractor, object, and coordinate detector and convert it to natural language. Then, the image caption, keywords, the required KPI, date, and marketer information is fed as input to the LLM and asked to output the image verbalization. This way, the LLM learns to map the image prompt and KPI to image verbalization.

We train the LLM on the train set of BoigBench and the stock data. In Listings~\ref{BoigLLM:verbalization-1}-\ref{BoigLLM:verbalization-2}, we provide the image caption, keywords, date, and required KPI as inputs to the model. Our aim is to train the model to predict the colors, tones, objects, and locations that should be reflected in the image. Moreover, we observe improved learning in the language model for behavior-conditioned image generation when introducing a 20\% noise in the behavior. We then task the model to rectify this noise in the output, simultaneously generating the verbalization of the behavior-conditioned image in Listings~\ref{BoigLLM:verbalization-3}-\ref{BoigLLM:verbalization-4}.

\begin{table*}[!htp]\centering
\caption{Performance of all models on the behavior-optimized image verbalization generation task across different KPI images in (i)~BoigBench data, (ii)~Stock data. It is noteworthy that (i)~BoigLLM outperforms larger sized GPT-3.5, 4 and also the sized Llama model fine-tuned on the same data (without including behavior tokens). (ii)~In-context learning does not work well in the behavior-conditioned image generation domain. Best results are given in \valbest{green} and runner-ups in \valgood{blue}.}\label{tab:twitter_boigbench_data_boig_llm_results}
\resizebox{0.8\textwidth}{!}{
\begin{tabular}{c|c|c|cccc|c|ccccc}\toprule
\multicolumn{12}{c}{\textbf{BoigBench Data}}\\\toprule
\multirow{2}{*}{\textbf{Model}}  & \multirow{2}{*}{\textbf{\makecell{Behaviour\\Optimised}}} &\multirow{2}{*}{\textbf{KPI}} &\multicolumn{4}{c|}{\textbf{Colours}} &\textbf{Tones} &\multicolumn{4}{c}{\textbf{Objects}} \\\cmidrule{4-12}
& & &\textbf{IOU} $\uparrow$ &\textbf{\makecell{Cosine\\Similarity}} $\uparrow$ &\textbf{\makecell{RGB\\distance}} $\downarrow$ &\textbf{\makecell{Coverage\\RMSE}} $\downarrow$ &\textbf{\makecell{Coverage\\RMSE}} $\downarrow$ &\textbf{IOU} $\uparrow$ &\textbf{\makecell{Cosine\\Similarity}} $\uparrow$ &\textbf{\makecell{Normalised\\Area RMSE}} $\downarrow$ &\textbf{\makecell{Relative\\Position Error}} $\downarrow$ \\\midrule
\multirow{2}{*}{\makecell{Finetuned\\Llama}} & \multirow{2}{*}{No} &High &\valgood{0.3717} &\valgood{0.8725} &0.2855 &\valgood{0.1694} &\valgood{0.1957} &\valgood{0.2547} &\valgood{0.8071} &\valgood{0.2612} &\valgood{0.3078} \\
&&Low &\valgood{0.3362} &\valgood{0.8602} &0.2223 &\valgood{0.1811} &\valgood{0.2339} &\valgood{0.2743} &\valgood{0.8047} &\valgood{0.2421} &\valgood{0.2954} \\
\midrule
\multirow{2}{*}{\makecell{Finetuned\\Llama (BoigLLM)}} & \multirow{2}{*}{\makecell{Behavior\\Finetuning}} &High &\valbest{0.4065} &\valbest{0.8898} &0.2795 &\valbest{0.1507} &\valbest{0.1718} &\valbest{0.2732} &\valbest{0.8122} &\valbest{0.2509} &\valbest{0.3054} \\
&&Low &\valbest{0.4531} &\valbest{0.8791} &\valbest{0.2084} &\valbest{0.1443} &\valbest{0.1848} &\valbest{0.3455} &\valbest{0.8228} &\valbest{0.2373} &\valbest{0.2889} \\\midrule
\multirow{2}{*}{3-shot GPT-3.5} &\multirow{2}{*}{\makecell{In-context\\learning}} &High &0.214 &0.7765 &0.2851 &0.1773 &0.396 &0.1085 &0.6621 &0.3090 &0.3651 \\
&&Low&0.2175 &0.7781 &0.2254 &0.2118 &0.3347 &0.1338 &0.6749 &0.3098 &0.3573 \\\midrule
\multirow{2}{*}{5-shot GPT-3.5} &\multirow{2}{*}{\makecell{In-context\\learning}} &High &0.2137 &0.7704 &0.2743 &0.1976 &0.324 &0.1011 &0.6456 &0.3160 &0.3622 \\
&&Low&0.2191 &0.7705 &\valgood{0.2176} &0.2449 &0.3186 &0.1264 &0.656 &0.3150 &0.3615 \\\midrule
\multirow{2}{*}{3-shot GPT-4} &\multirow{2}{*}{\makecell{In-context\\learning}} &High  &0.2421 &0.7887 &\valgood{0.2726} &0.192 &0.304 &0.1035 &0.6316 &0.3137 &0.3666 \\
&&Low&0.2405 &0.793 &0.2332 &0.2094 &0.3037 &0.1419 &0.6604 &0.3248 &0.3763 \\\midrule
\multirow{2}{*}{5-shot GPT-4} &\multirow{2}{*}{\makecell{In-context\\learning}} &High &0.2437 &0.7905 &\valbest{0.2702} &0.1864 &0.2937 &0.1008 &0.6136 &0.3111 &0.3782 \\
&&Low&0.2448 &0.7924 &0.2278 &0.2144 &0.2944 &0.1464 &0.6406 &0.3301 &0.3857 \\
\bottomrule
\end{tabular}}

\resizebox{0.8\textwidth}{!}{
\begin{tabular}{c|c|c|cccc|c|ccccc}\toprule
\multicolumn{12}{c}{\textbf{Stock Data}}\\\toprule
\multirow{2}{*}{\textbf{Model}}  & \multirow{2}{*}{\textbf{\makecell{Behaviour\\Optimised}}} &\multirow{2}{*}{\textbf{KPI}} &\multicolumn{4}{c|}{\textbf{Colours}} &\textbf{Tones} &\multicolumn{4}{c}{\textbf{Objects}} \\\cmidrule{4-12}
& & &\textbf{IOU} $\uparrow$ &\textbf{\makecell{Cosine\\Similarity}} $\uparrow$ &\textbf{\makecell{RGB\\distance}} $\downarrow$ &\textbf{\makecell{Coverage\\RMSE}} $\downarrow$ &\textbf{\makecell{Coverage\\RMSE}} $\downarrow$ &\textbf{IOU} $\uparrow$ &\textbf{\makecell{Cosine\\Similarity}} $\uparrow$ &\textbf{\makecell{Normalised\\Area RMSE}} $\downarrow$ &\textbf{\makecell{Relative\\Position Error}} $\downarrow$ \\\midrule
\multirow{3}{*}{\makecell{Finetuned\\Llama}} & \multirow{3}{*}{No} &High &\valgood{0.3591} &\valgood{0.8622} &\valgood{0.2134} &\valgood{0.1830} &\valbest{0.1893} &\valgood{0.2804} &\valgood{0.7000} &\valgood{0.2475} &\valbest{0.2913} \\
& &Medium &\valgood{0.3404} &\valgood{0.8569} &\valgood{0.2219} &\valgood{0.1837} &\valgood{0.1845} &\valgood{0.2641} &\valgood{0.6716} &\valgood{0.2619} &\valbest{0.3000} \\
& &Low &\valgood{0.3373} &\valgood{0.862} &\valgood{0.2296} &\valgood{0.1792} &\valgood{0.2002} &\valgood{0.241} &\valgood{0.6758} &\valgood{0.2522} &\valbest{0.3085} \\
\midrule
\multirow{3}{*}{\makecell{Finetuned\\Llama (BoigLLM)}} &\multirow{3}{*}{\makecell{Behavior\\Finetuning}} &High &\valbest{0.3684} &\valbest{0.8632} &\valbest{0.2021} &\valbest{0.1777} &\valgood{0.1963} &\valbest{0.3046} &\valbest{0.7016} &\valbest{0.2419} &\valgood{0.3041} \\
& &Medium &\valbest{0.366} &\valbest{0.8651} &\valbest{0.2135} &\valbest{0.1813} &\valbest{0.1752} &\valbest{0.273} &\valbest{0.6729} &\valbest{0.2520} &\valgood{0.3103} \\
& &Low &\valbest{0.3514} &\valbest{0.8722} &\valbest{0.2233} &\valbest{0.1655} &\valbest{0.1989} &\valbest{0.2499} &\valbest{0.6687} &\valbest{0.2466} &\valgood{0.3217} \\\midrule
\multirow{3}{*}{3-shot GPT-3.5} &\multirow{3}{*}{\makecell{In-context\\learning}} &High &0.2205 &0.8126 &0.2516 &0.2734 &0.3094 &0.0974 &0.4965 &0.3724 &0.4183 \\
&&Medium &0.2242 &0.8069 &0.2607 &0.2542 &0.3103 &0.0865 &0.4891 &1.1761 &0.4287 \\
&&Low &0.2166 &0.8118 &0.2628 &0.2494 &0.3104 &0.1005 &0.5037 &0.3630 &0.4070 \\\midrule
\multirow{3}{*}{5-shot GPT-3.5} &\multirow{3}{*}{\makecell{In-context\\learning}} &High &0.2354 &0.8208 &0.2579 &0.258 &0.3262 &0.1063 &0.5454 &0.3589 &0.4086 \\
&&Medium &0.2275 &0.8078 &0.2667 &0.2365 &0.3184 &0.0942 &0.5024 &0.4252 &0.4149 \\
&&Low &0.2251 &0.8146 &0.2672 &0.2322 &0.3207 &0.1074 &0.5179 &0.3511 &0.4045 \\\midrule
\multirow{3}{*}{3-shot GPT-4} &\multirow{3}{*}{\makecell{In-context\\learning}} &High &0.2721 &0.8103 &0.2193 &0.2282 &0.2953 &0.0961 &0.5239 &0.3477 &0.4072 \\
&&Medium &0.2686 &0.8105 &0.2363 &0.2147 &0.2751 &0.106 &0.5092 &0.4748 &0.4183 \\
&&Low &0.2425 &0.814 &0.245 &0.2145 &0.3002 &0.1012 &0.5269 &0.3362 &0.4131 \\\midrule
\multirow{3}{*}{5-shot GPT-4} &\multirow{3}{*}{\makecell{In-context\\learning}} &High &0.289 &0.819 &0.2209 &0.2244 &0.3004 &0.1118 &0.5558 &0.3481 &0.3929 \\
&&Medium &0.2752 &0.812 &0.2399 &0.215 &0.2761 &0.1151 &0.5405 &0.4573 &0.3990 \\
&&Low &0.2481 &0.8076 &0.251 &0.2086 &0.2938 &0.1027 &0.5428 &0.3305 &0.3965 \\
\bottomrule
\end{tabular}}
\end{table*}

\subsection{Performance Aligning Stable Diffusion}
\label{sec:Performance Aligning Stable Diffusion}
We use BoigLLM as a reward model in order to align a trained stable diffusion image generation model to generate more performant images. The entire process of alignment is shown in Figure~\ref{fig:architecture_diagram}. The policy gradient feedback loop shown in the figure for aligning the stable diffusion model is done using the denoising diffusion policy optimization (DDPO)~\cite{black2023training} algorithm. The denoising process in diffusion models is a multi-step recursive process with a pre-specified finite number of steps. In DDPO, this denoising process is viewed as a finite horizon Markov decision process (MDP), where the state comprises of the current context, number of steps left in the process and the current denoised image. The action to be taken is to predict the next image using this state. The image forming the initial state is sampled from a standard normal distribution. Mathematically, a finite horizon MDP is defined as a tuple $\{T, \mathcal{S}, \mathcal{A}, P, R\}$, where these components are defined as
\begin{enumerate}
    \item $T$ is the horizon or the number of steps in the MDP
    \item $\mathcal{S}$ is the state space. Here it comprises of three components, the context $c$, the current number of steps left in the denoising process, $t$, and the current denoised image representation (a given vector encoding of the image), $x_t$. The initial or starting state has the context $c_0$ given as input, the number of steps left at the beginning, $t_0 = T$ and the initial image representation is sampled from a normal distribution of appropriate dimension, $x_0 \sim \mathcal{N}(0, I)$.
    \item $\mathcal{A}$ is the action space, and here it is the space comprising of all image representations $x$. If $x$ is a $d-$dimensional vector, then $\mathcal{A} = \mathbb{R}^d$.
    \item $P: \mathcal{S} \times \mathcal{A} \to \Delta(\mathcal{S})$ is the transition function. Here, we specify $P$ separately for each of the three components of the state as $P_c = \delta(c_t)$, $P_t = \delta(t-1)$, and $P_x = \delta(a_t)$, where the current state is $c_t, t, x_t$, current action $a_t = x_{t-1}$, and $\delta(\cdot)$ is the Dirac delta distribution.
    \item $R: \mathcal{S} \times \mathcal{A} \to \mathbb{R}$ is the reward function that takes a state and action as input \cy{better to specify what the state and action contain in our setting} and returns a scalar reward. We generate this scalar reward signal using BoigLLM.
\end{enumerate}

 \noindent The process of constructing the reward function comprises of the following steps (Fig.~\ref{fig:architecture_diagram}):
\begin{enumerate}
    \item We featurize the generated image to yield features that BoigLLM is meant to predict (including colors, tones, objects, and their positions).
    
    \item We use BoigLLM to predict the logits of the text comprising of a high KPI label, the textual description that served as input for the stable diffusion model along with the verbalized features of the image generated by stable diffusion as described in the previous step.
    \item We now have one logit per text token as BoigLLM's output. To convert this to a scalar score, we compute the probabilities of each token and then add them\footnote{We tried other variants such as computing the probability of the text as a whole by adding the log probabilities instead of the probabilties, thresholding these scores to ensure that they remain bounded, and also using affine transformations on the sum of probabilities as well as sum of log probabilities. However, we found that empirically the sum of probabilities score works best as a reward model for our purpose in terms of the observed FID and aesthetic scores.}.
    
\end{enumerate}

DDPO is based on conventional policy gradient algorithms in RL, which assumes that the environment, \textit{i.e.}, the transition and the reward functions are neither differentiable nor accessible to the RL agent. However, when one has access to both the transition and reward functions, and when both these are differentiable, one could use analytic policy gradient based algorithms for training the RL agent~\cite{wiedemann2023training}. In our case, we do have access to the transition and reward functions. As in standard diffusion models, the transition function is the single-step denoising function. However, we do not have a differentiable reward function as we are using non-differentiable featurizers in Step 1 of the reward generation process as described above. An alternative approach would be to make this step differentiable and then use the end-to-end analytic policy gradient approach for aligning the stable diffusion model. In order to simplify our training pipeline and to avoid getting into potential stability issues when performing end-to-end learning, we chose the conventional RL approach of DDPO for this work.%

\subsection{Finetuning Stable Diffusion on High-KPI Images}
\label{sec:Finetuning Stable Diffusion on High-KPI Images}

Several prior studies have demonstrated the feasibility of learning styles through stable diffusion by fine-tuning the model \cite{vonplaten2023stablediffusion,pinkney2022text-to-pokemon,cjwbw2022waifudiffusion,prompthero2023openjourneyv2,everaert2023diffusion}. These approaches typically involve fine-tuning the U-Net architecture within the Stable Diffusion framework using a set of images exhibiting the desired style.
In an attempt to determine whether learning high Key Performance Indicators (KPIs) follows a similar approach to learning image styles, we performed fine-tuning on the base Stable Diffusion U-Net using the training set, following the procedure outlined by \cite{vonplaten2023stablediffusion}. We fine-tuned the base model for $50$ epochs using the high KPI samples from both the BoigBench and Stock datasets. However, our observations reveal that fine-tuning the stable diffusion model does not yield improvements in rewards calculated over the validation set across subsequent epochs.

\section{Evaluation and Discussion}
\label{sec:evaluation and discussion}

\textbf{Behavior-conditioned Image Verbalization Generation:}
To generate behavior-conditioned image verbalization, we compare several models: in-context trained GPT-3.5 and GPT-4, behavior-finetuned Llama (BoigLLM), and Llama fine-tuned on image verbalization but without user behavior information. By comparing against a fine-tuned Llama trained on the same instruction as BoigLLM, except with the inclusion of behavior tokens, we aim to isolate the impact of behavior tokens on improving generated behavior-conditioned image verbalizations, independent of the instruction tuning process. We assess all models across multiple metrics that evaluate the extent to which the generated verbalizations align with ground truth in terms of colors, tones, objects, and their positions. Intersection over Union (IoU) metrics gauge the overlap between ground truth and generated constructs (colors and objects), while similarity metrics measure cosine similarity between ground truth and generated constructs (colors, objects). Coverage errors determine the how closely the proportion of ground truth and predicted constructs (colors, tones) in the image match. Additionally, we calculate differences in predicted and ground truth areas and locations for objects, accounting for semantically similar objects (such as sofa and couch). Further details on these metrics and their formulas can be found in Appendix~\ref{sec:Evaluation Metrics}.

Table~\ref{tab:twitter_boigbench_data_boig_llm_results} displays the outcomes. The results indicate that behavior fine-tuning enables BoigLLM to achieve superior performance across all metrics, surpassing both equivalently sized fine-tuned Llama and 10x larger instruction-tuned GPT-3.5 and GPT-4. Notably, Medium and Low KPI test samples exhibit more substantial improvements compared to the High KPI samples, despite the latter having a slightly higher quantity. Furthermore, in-context learning demonstrates subpar performance, with both the three and five-shot models displaying similar results.

\textbf{Behavior-optimized image generation:}
Next, we compare three models for the task of behavior-optimized image generation: base stable diffusion, stable diffusion finetuned on high-KPI samples, and performance-aligned stable diffusion (BoigSD). Consistent with prior RLHF literature \cite{black2023training,kirstain2023pick,xu2023imagereward,fan2023dpok,wu2023better}, we evaluate these models on rewards provided by BoigLLM. We also include other metrics like FID \cite{heusel2017gans}, aesthetics \cite{schuhmann2022laion}, and CLIP score \cite{radford2021learning}. 
The results are summarized in Table~\ref{tab:twitter_data_scores}. We see that BoigSD outperforms base stable diffusion and high-KPI finetuned stable diffusion. Intriguingly, further finetuning the stable diffusion on high-KPI samples results in a decreased mean reward. Figs.~\ref{fig:journey of boigSD images},\ref{img:comparison-marketing-image} visually depict the progression of several images during BoigSD training.

Next, we evaluate the side effects of BoigLLM's reward on BoigSD training. We notice (Table~\ref{tab:twitter_data_scores}) that as a result of BoigLLM reward, in spite of not being directly targeted, aesthetics gets optimized across all KPI buckets for both datasets. This could be a side effect of optimizing on marketing images, which are often of higher aesthetics than images sourced from other image datasets. Further, as a consequence of training on marketing images, we find that stable diffusion learns to generate images that are more aligned with marketing. For instance, Fig.~\ref{img:comparison-marketing-image} shows several examples of product, model, and travel photography generated by BoigSD and base SD, demonstrating BoigSD's biases towards certain persuasion strategies such as social appeal and social identity, commonly observed in marketing scenarios \cite{singla2022persuasion} but not found in general photography.

\begin{table}[!htp]\centering
\caption{Results for the performance of various models on the BoigBench and Stock datasets for the behavior-optimized image generation task. Results are computed on the BoigLLM reward as well as other metrics reported in the literature. BoigLLM aligned Stable Diffusion (BoigSD) achieves the highest reward. Results on other metrics are mixed, showing that while stable diffusion is inherently trained for metrics like image aesthetics and fidelity, behavior is a different metric and needs separate training. BoigSD, while trained on behavior tokens, does not lose out on other metrics but gains on the behavior-based reward. We see that supervised fine-tuning of Stable Diffusion on high-KPI samples does not help it in learning behavior. Best results are given in \valbest{green} and runner-ups in \valgood{blue}.\label{tab:twitter_data_scores}}
\resizebox{0.5\textwidth}{!}{
\begin{tabular}{ccc|ccc}\toprule%
\multicolumn{6}{c}{\textbf{BoigBench}}\\\midrule
\multirow{2}{*}{\textbf{Images}} &\multirow{2}{*}{\textbf{KPI}} &\multirow{2}{*}{\textbf{Reward} $\uparrow$} & \multicolumn{3}{c}{\textbf{Other Metrics}}\\
&&&\textbf{FID} $\downarrow$ & \textbf{Aesthetic Score} $\uparrow$ &\textbf{CLIP Score} $\uparrow$ \\\midrule%
\multirow{2}{*}{Base Stable Diffusion} & High & \valgood{242.545} & \valgood{34.958} & \valgood{5.221} & \valgood{33.346} \\
&Low & \valgood{238.471} & \valgood{42.999} & \valgood{4.925} & \valgood{31.705} \\
\midrule
\multirow{2}{*}{\makecell{High-KPI fine-tuned\\Stable Diffusion}} &High & 239.023 & \valbest{26.023} & 4.850  & 32.210 \\
&Low & 223.619 & \valbest{37.497} & 4.433 & 30.979 \\
\midrule
\multirow{2}{*}{\makecell{BoigLLM aligned\\Stable Diffusion (BoigSD)}} &High & \valbest{254.918} & 36.546 & \valbest{5.341} & \valbest{33.379} \\
&Low & \valbest{247.597} & 49.492 & \valbest{5.087} & \valbest{31.719} \\\bottomrule%
\end{tabular}}

\resizebox{0.5\textwidth}{!}{
\begin{tabular}{ccc|ccc}\toprule%
\multicolumn{6}{c}{\textbf{Stock Data}}\\\midrule
\multirow{2}{*}{\textbf{Images}} &\multirow{2}{*}{\textbf{KPI}} &\multirow{2}{*}{\textbf{Reward} $\uparrow$} & \multicolumn{3}{c}{\textbf{Other Metrics}}\\
&&&\textbf{FID} $\downarrow$ & \textbf{Aesthetic Score} $\uparrow$ &\textbf{CLIP Score} $\uparrow$\\\midrule%
\multirow{3}{*}{Base Stable Diffusion} & High & \valgood{202.459} & \valgood{33.481} & \valgood{5.256} & \valbest{30.659} \\
&Medium & \valgood{203.922} & \valgood{30.936} & 5.269 & \valbest{30.442} \\
&Low & \valgood{203.328} & \valgood{29.072} &5.279 & \valbest{30.293}\\
\midrule
\multirow{3}{*}{\makecell{High-KPI fine-tuned\\Stable Diffusion}} &High & 190.153 & \valbest{33.217} & 5.248 & 30.347\\
&Medium & 195.033 & \valbest{30.056} & \valgood{5.304} & 30.115\\
&Low & 195.541 & \valbest{28.893}& \valgood{5.321} & 29.975\\
\midrule
\multirow{3}{*}{\makecell{BoigLLM aligned\\Stable Diffusion\\(BoigSD)}} &High & \valbest{226.131} & 35.406 & \valbest{5.382} &\valgood{30.616} \\
&Medium & \valbest{224.571} & 32.376 & \valbest{5.406} &\valgood{30.395}\\
&Low & \valbest{225.511} & 30.367 & \valbest{5.417} & \valgood{30.288}\\\midrule%
\end{tabular}}

\end{table}

\section{Conclusion}
Image generation technologies have undergone a significant evolution, transitioning from research concepts to viable commercial products. The initial phase of image generation primarily focused on producing higher-quality images that adhere to provided prompts. In the next phase, generated images should not only meet quality standards but also align with the creator's objectives. This necessitates conditioning the image generation process based on the utility of the generated image. In marketing contexts, this utility translates into achieving higher customer engagement metrics such as likes, shares, clicks, and more. In this paper, we introduce the problem statement of behavior-conditioned image generation and propose the first large-scale dataset for this purpose. Additionally, we present the results of several techniques to solve this problem both in the natural language domain by generating behavior-conditioned image generation and in the computer vision space by generating actual pixels. Moreover, we demonstrate that current models, due to their lack of conditioning on behavior tokens, struggle to produce behavior-optimized images.

{%
\small
\bibliographystyle{ieee_fullname}
\bibliography{egbib}

\begin{thebibliography}{10}\itemsep=-1pt

\bibitem{alalwan2017social}
Ali~Abdallah Alalwan, Nripendra~P Rana, Yogesh~K Dwivedi, and Raed Algharabat.
\newblock Social media in marketing: A review and analysis of the existing literature.
\newblock {\em Telematics and informatics}, 34(7):1177--1190, 2017.

\bibitem{bhat2020near}
Nikhil Bhat, Vivek~F Farias, Ciamac~C Moallemi, and Deeksha Sinha.
\newblock Near-optimal ab testing.
\newblock {\em Management Science}, 66(10):4477--4495, 2020.

\bibitem{bhattacharya2023video}
Aanisha Bhattacharya, Yaman~K Singla, Balaji Krishnamurthy, Rajiv~Ratn Shah, and Changyou Chen.
\newblock A video is worth 4096 tokens: Verbalize story videos to understand them in zero shot.
\newblock {\em arXiv preprint arXiv:2305.09758}, 2023.

\bibitem{black2023training}
Kevin Black, Michael Janner, Yilun Du, Ilya Kostrikov, and Sergey Levine.
\newblock Training diffusion models with reinforcement learning.
\newblock {\em arXiv preprint arXiv:2305.13301}, 2023.

\bibitem{bobadilla2013recommender}
Jes{\'u}s Bobadilla, Fernando Ortega, Antonio Hernando, and Abraham Guti{\'e}rrez.
\newblock Recommender systems survey.
\newblock {\em Knowledge-based systems}, 46:109--132, 2013.

\bibitem{christiano2017deep}
Paul~F Christiano, Jan Leike, Tom Brown, Miljan Martic, Shane Legg, and Dario Amodei.
\newblock Deep reinforcement learning from human preferences.
\newblock {\em Advances in neural information processing systems}, 30, 2017.

\bibitem{cjwbw2022waifudiffusion}
Cjwbw.
\newblock Waifudiffusion.
\newblock \url{https://replicate.com/cjwbw/waifu-diffusion}, 2022.

\bibitem{clark2023directly}
Kevin Clark, Paul Vicol, Kevin Swersky, and David~J Fleet.
\newblock Directly fine-tuning diffusion models on differentiable rewards.
\newblock {\em arXiv preprint arXiv:2309.17400}, 2023.

\bibitem{dhariwal2021diffusion}
Prafulla Dhariwal and Alexander Nichol.
\newblock Diffusion models beat gans on image synthesis.
\newblock {\em Advances in neural information processing systems}, 34:8780--8794, 2021.

\bibitem{dosovitskiy2020image}
Alexey Dosovitskiy, Lucas Beyer, Alexander Kolesnikov, Dirk Weissenborn, Xiaohua Zhai, Thomas Unterthiner, Mostafa Dehghani, Matthias Minderer, Georg Heigold, Sylvain Gelly, et~al.
\newblock An image is worth 16x16 words: Transformers for image recognition at scale.
\newblock {\em arXiv preprint arXiv:2010.11929}, 2020.

\bibitem{everaert2023diffusion}
Martin~Nicolas Everaert, Marco Bocchio, Sami Arpa, Sabine S{\"u}sstrunk, and Radhakrishna Achanta.
\newblock Diffusion in style.
\newblock In {\em Proceedings of the IEEE/CVF International Conference on Computer Vision}, pages 2251--2261, 2023.

\bibitem{fan2023dpok}
Ying Fan, Olivia Watkins, Yuqing Du, Hao Liu, Moonkyung Ryu, Craig Boutilier, Pieter Abbeel, Mohammad Ghavamzadeh, Kangwook Lee, and Kimin Lee.
\newblock Dpok: Reinforcement learning for fine-tuning text-to-image diffusion models.
\newblock {\em arXiv preprint arXiv:2305.16381}, 2023.

\bibitem{heusel2017gans}
Martin Heusel, Hubert Ramsauer, Thomas Unterthiner, Bernhard Nessler, and Sepp Hochreiter.
\newblock Gans trained by a two time-scale update rule converge to a local nash equilibrium.
\newblock {\em Advances in neural information processing systems}, 30, 2017.

\bibitem{ho2020denoising}
Jonathan Ho, Ajay Jain, and Pieter Abbeel.
\newblock Denoising diffusion probabilistic models.
\newblock {\em Advances in neural information processing systems}, 33:6840--6851, 2020.

\bibitem{khandelwal2023large}
Ashmit Khandelwal, Aditya Agrawal, Aanisha Bhattacharyya, Yaman~K Singla, Somesh Singh, Uttaran Bhattacharya, Ishita Dasgupta, Stefano Petrangeli, Rajiv~Ratn Shah, Changyou Chen, et~al.
\newblock Large content and behavior models to understand, simulate, and optimize content and behavior.
\newblock {\em arXiv preprint arXiv:2309.00359}, 2023.

\bibitem{kirstain2023pick}
Yuval Kirstain, Adam Polyak, Uriel Singer, Shahbuland Matiana, Joe Penna, and Omer Levy.
\newblock Pick-a-pic: An open dataset of user preferences for text-to-image generation.
\newblock {\em arXiv preprint arXiv:2305.01569}, 2023.

\bibitem{kreutzer2018can}
Julia Kreutzer, Shahram Khadivi, Evgeny Matusov, and Stefan Riezler.
\newblock Can neural machine translation be improved with user feedback?
\newblock {\em arXiv preprint arXiv:1804.05958}, 2018.

\bibitem{singla2022persuasion}
Yaman Kumar, Rajat Jha, Arunim Gupta, Milan Aggarwal, Aditya Garg, Tushar Malyan, Ayush Bhardwaj, Rajiv~Ratn Shah, Balaji Krishnamurthy, and Changyou Chen.
\newblock Persuasion strategies in advertisements.
\newblock {\em Proceedings of the AAAI Conference on Artificial Intelligence}, 2023.

\bibitem{lasswell1948structure}
Harold~D Lasswell.
\newblock The structure and function of communication in society.
\newblock {\em The communication of ideas}, 37(1):136--139, 1948.

\bibitem{lee2023aligning}
Kimin Lee, Hao Liu, Moonkyung Ryu, Olivia Watkins, Yuqing Du, Craig Boutilier, Pieter Abbeel, Mohammad Ghavamzadeh, and Shixiang~Shane Gu.
\newblock Aligning text-to-image models using human feedback.
\newblock {\em arXiv preprint arXiv:2302.12192}, 2023.

\bibitem{li2023blip2}
Junnan Li, Dongxu Li, Silvio Savarese, and Steven Hoi.
\newblock Blip-2: Bootstrapping language-image pre-training with frozen image encoders and large language models, 2023.

\bibitem{liu2023visual}
Haotian Liu, Chunyuan Li, Qingyang Wu, and Yong~Jae Lee.
\newblock Visual instruction tuning.
\newblock {\em arXiv preprint arXiv:2304.08485}, 2023.

\bibitem{liu2023grounding}
Shilong Liu, Zhaoyang Zeng, Tianhe Ren, Feng Li, Hao Zhang, Jie Yang, Chunyuan Li, Jianwei Yang, Hang Su, Jun Zhu, et~al.
\newblock Grounding dino: Marrying dino with grounded pre-training for open-set object detection.
\newblock {\em arXiv preprint arXiv:2303.05499}, 2023.

\bibitem{lu2023specialist}
Haoming Lu, Hazarapet Tunanyan, Kai Wang, Shant Navasardyan, Zhangyang Wang, and Humphrey Shi.
\newblock Specialist diffusion: Plug-and-play sample-efficient fine-tuning of text-to-image diffusion models to learn any unseen style.
\newblock In {\em Proceedings of the IEEE/CVF Conference on Computer Vision and Pattern Recognition}, pages 14267--14276, 2023.

\bibitem{menick2022teaching}
Jacob Menick, Maja Trebacz, Vladimir Mikulik, John Aslanides, Francis Song, Martin Chadwick, Mia Glaese, Susannah Young, Lucy Campbell-Gillingham, Geoffrey Irving, et~al.
\newblock Teaching language models to support answers with verified quotes.
\newblock {\em arXiv preprint arXiv:2203.11147}, 2022.

\bibitem{min-etal-2022-rethinking}
Sewon Min, Xinxi Lyu, Ari Holtzman, Mikel Artetxe, Mike Lewis, Hannaneh Hajishirzi, and Luke Zettlemoyer.
\newblock Rethinking the role of demonstrations: What makes in-context learning work?
\newblock In Yoav Goldberg, Zornitsa Kozareva, and Yue Zhang, editors, {\em Proceedings of the 2022 Conference on Empirical Methods in Natural Language Processing}, pages 11048--11064, Abu Dhabi, United Arab Emirates, Dec. 2022. Association for Computational Linguistics.

\bibitem{nadiger2019federated}
Chetan Nadiger, Anil Kumar, and Sherine Abdelhak.
\newblock Federated reinforcement learning for fast personalization.
\newblock In {\em 2019 IEEE Second International Conference on Artificial Intelligence and Knowledge Engineering (AIKE)}, pages 123--127. IEEE, 2019.

\bibitem{nakano2021webgpt}
Reiichiro Nakano, Jacob Hilton, Suchir Balaji, Jeff Wu, Long Ouyang, Christina Kim, Christopher Hesse, Shantanu Jain, Vineet Kosaraju, William Saunders, et~al.
\newblock Webgpt: Browser-assisted question-answering with human feedback.
\newblock {\em arXiv preprint arXiv:2112.09332}, 2021.

\bibitem{openai2023gpt4}
OpenAI.
\newblock Gpt-4 technical report, 2023.

\bibitem{ouyang2022training}
Long Ouyang, Jeffrey Wu, Xu Jiang, Diogo Almeida, Carroll Wainwright, Pamela Mishkin, Chong Zhang, Sandhini Agarwal, Katarina Slama, Alex Ray, et~al.
\newblock Training language models to follow instructions with human feedback.
\newblock {\em Advances in Neural Information Processing Systems}, 35:27730--27744, 2022.

\bibitem{pinkney2022text-to-pokemon}
Justin Pinkney.
\newblock Text-to-pokemon.
\newblock \url{https://replicate.com/lambdal/text-to-pokemon}, 2022.

\bibitem{pressman2023simulacra}
John~David Pressman, Katherine Crowson, and Simulacra~Captions Contributors.
\newblock Simulacra aesthetic captions, 2023.

\bibitem{prompthero2023openjourneyv2}
PromptHero.
\newblock Openjourneyv2.
\newblock \url{https://huggingface.co/prompthero/openjourney-v2}, 2023.

\bibitem{Qin_2020_PR}
Xuebin Qin, Zichen Zhang, Chenyang Huang, Masood Dehghan, Osmar Zaiane, and Martin Jagersand.
\newblock U2-net: Going deeper with nested u-structure for salient object detection.
\newblock volume 106, page 107404, 2020.

\bibitem{radford2021learning}
Alec Radford, Jong~Wook Kim, Chris Hallacy, Aditya Ramesh, Gabriel Goh, Sandhini Agarwal, Girish Sastry, Amanda Askell, Pamela Mishkin, Jack Clark, et~al.
\newblock Learning transferable visual models from natural language supervision.
\newblock In {\em International conference on machine learning}, pages 8748--8763. PMLR, 2021.

\bibitem{rombach2022high}
Robin Rombach, Andreas Blattmann, Dominik Lorenz, Patrick Esser, and Bj{\"o}rn Ommer.
\newblock High-resolution image synthesis with latent diffusion models.
\newblock In {\em Proceedings of the IEEE/CVF conference on computer vision and pattern recognition}, pages 10684--10695, 2022.

\bibitem{ruiz2023dreambooth}
Nataniel Ruiz, Yuanzhen Li, Varun Jampani, Yael Pritch, Michael Rubinstein, and Kfir Aberman.
\newblock Dreambooth: Fine tuning text-to-image diffusion models for subject-driven generation.
\newblock In {\em Proceedings of the IEEE/CVF Conference on Computer Vision and Pattern Recognition}, pages 22500--22510, 2023.

\bibitem{saharia2022photorealistic}
Chitwan Saharia, William Chan, Saurabh Saxena, Lala Li, Jay Whang, Emily~L Denton, Kamyar Ghasemipour, Raphael Gontijo~Lopes, Burcu Karagol~Ayan, Tim Salimans, et~al.
\newblock Photorealistic text-to-image diffusion models with deep language understanding.
\newblock {\em Advances in Neural Information Processing Systems}, 35:36479--36494, 2022.

\bibitem{schuhmann2022laion}
Christoph Schuhmann, Romain Beaumont, Richard Vencu, Cade Gordon, Ross Wightman, Mehdi Cherti, Theo Coombes, Aarush Katta, Clayton Mullis, Mitchell Wortsman, et~al.
\newblock Laion-5b: An open large-scale dataset for training next generation image-text models.
\newblock {\em Advances in Neural Information Processing Systems}, 35:25278--25294, 2022.

\bibitem{shannon-weaver-1949}
Claude~E. Shannon and Warren Weaver.
\newblock {\em The mathematical theory of communication.}
\newblock The mathematical theory of communication. University of Illinois Press, Champaign, IL, US, 1949.

\bibitem{stiennon2020learning}
Nisan Stiennon, Long Ouyang, Jeffrey Wu, Daniel Ziegler, Ryan Lowe, Chelsea Voss, Alec Radford, Dario Amodei, and Paul~F Christiano.
\newblock Learning to summarize with human feedback.
\newblock {\em Advances in Neural Information Processing Systems}, 33:3008--3021, 2020.

\bibitem{touvron2023llama2}
Hugo Touvron, Louis Martin, Kevin Stone, Peter Albert, Amjad Almahairi, Yasmine Babaei, Nikolay Bashlykov, Soumya Batra, Prajjwal Bhargava, Shruti Bhosale, et~al.
\newblock Llama 2: Open foundation and fine-tuned chat models.
\newblock {\em arXiv preprint arXiv:2307.09288}, 2023.

\bibitem{vonplaten2023stablediffusion}
Patrick von Platen, Suraj Patil, Anton Lozhkov, Pedro Cuenca, Nathan Lambert, Kashif Rasul, Mishig Davaadorj, and Thomas Wolf.
\newblock Stablediffusion text-to-image fine-tuning — huggingface diffusers documentation.
\newblock \url{https://huggingface.co/docs/diffusers/v0.13.0/en/training/text2image}, 2023.

\bibitem{wiedemann2023training}
Nina Wiedemann, Valentin Wüest, Antonio Loquercio, Matthias Müller, Dario Floreano, and Davide Scaramuzza.
\newblock Training efficient controllers via analytic policy gradient, 2023.

\bibitem{wikidata}
{Wikidata contributors}.
\newblock {Wikidata}.
\newblock \url{https://www.wikidata.org/}, ongoing.

\bibitem{wu2023better}
Xiaoshi Wu, Keqiang Sun, Feng Zhu, Rui Zhao, and Hongsheng Li.
\newblock Better aligning text-to-image models with human preference.
\newblock {\em arXiv preprint arXiv:2303.14420}, 2023.

\bibitem{xu2023imagereward}
Jiazheng Xu, Xiao Liu, Yuchen Wu, Yuxuan Tong, Qinkai Li, Ming Ding, Jie Tang, and Yuxiao Dong.
\newblock Imagereward: Learning and evaluating human preferences for text-to-image generation.
\newblock {\em arXiv preprint arXiv:2304.05977}, 2023.

\bibitem{zhang2023adding}
Lvmin Zhang, Anyi Rao, and Maneesh Agrawala.
\newblock Adding conditional control to text-to-image diffusion models.
\newblock In {\em Proceedings of the IEEE/CVF International Conference on Computer Vision}, pages 3836--3847, 2023.

\bibitem{zhang2023recognize}
Youcai Zhang, Xinyu Huang, Jinyu Ma, Zhaoyang Li, Zhaochuan Luo, Yanchun Xie, Yuzhuo Qin, Tong Luo, Yaqian Li, Shilong Liu, et~al.
\newblock Recognize anything: A strong image tagging model.
\newblock {\em arXiv preprint arXiv:2306.03514}, 2023.

\bibitem{zhu2023minigpt}
Deyao Zhu, Jun Chen, Xiaoqian Shen, Xiang Li, and Mohamed Elhoseiny.
\newblock Minigpt-4: Enhancing vision-language understanding with advanced large language models.
\newblock {\em arXiv preprint arXiv:2304.10592}, 2023.

\bibitem{ziegler2019fine}
Daniel~M Ziegler, Nisan Stiennon, Jeffrey Wu, Tom~B Brown, Alec Radford, Dario Amodei, Paul Christiano, and Geoffrey Irving.
\newblock Fine-tuning language models from human preferences.
\newblock {\em arXiv preprint arXiv:1909.08593}, 2019.

\end{thebibliography}
}

\clearpage
\appendix
\onecolumn 
\section*{Appendix}

\begin{figure}
  \centering
  \includegraphics[width=1.0\columnwidth]{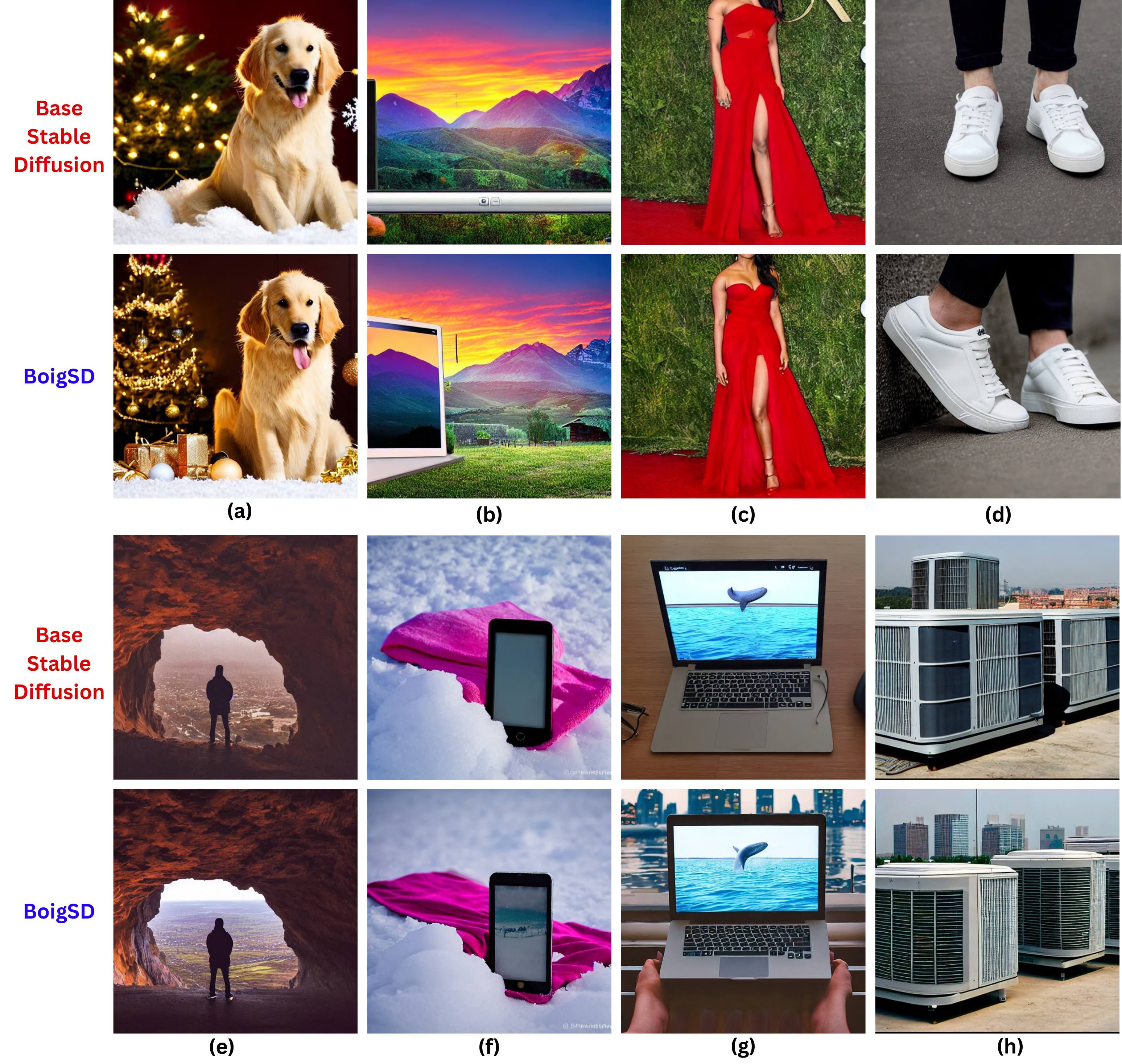}
  \caption{Comparison of generated images - BoigSD \textit{vs} Base stable diffusion. BoigSD learns temporal patterns (a) (Christmas themed image of dog), generates better product photography (b,f,g,h), model photography (c,d), travel photography (e), and generates images with social appeal and social identity (f,g,h). \label{img:comparison-marketing-image}}
\end{figure}

\section{Training BOIG-LLM}
\label{sec:Training BOIG-LLM}

\subsection{Evaluation Metrics}
\label{sec:Evaluation Metrics}
\begin{itemize}
    \item \textbf{Colours IOU}: The intersection over union between set $C^G$ of colours in the ground truth image verbalization and set $C^P$ of colours in the predicted image verbalization is computed as:
    \begin{equation}
        IOU(C^G, C^P) = \frac{|C^G \cap C^P|}{|C^G \cup C^P|}
    \end{equation}
    \item \textbf{Colours similarity}: For the ground truth colour set $C^G = {c_1^G, c_2^G, ..., c_i^G}$ and predicted colour set $C^P = \{c_1^P, c_2^P, ..., c_j^P\}$, we correspondingly obtain the sets of word vectors $W^G = \{w_1^G, w_2^G, ..., w_i^G\}$ and $W^P = \{w_1^P, w_2^P, ..., w_j^P\}$, using Spacy \footnote{\url{https://spacy.io/}}. For some similarity threshold $\tau$, the mean cosine similarity is computed as follows:
    \begin{equation}
        \frac{\mathop{\sum_{i=1}^{|C^G|}\sum_{j=1}^{|C^P|}} cos(w_i^G, w_j^P).I(w_i^G, w_j^P, \tau)}{\mathop{\sum_{i=1}^{|C^G|}\sum_{j=1}^{|C^P|}} I(w_i^G, w_j^P, \tau) }
    \end{equation}
    where $I(w_i^G, w_j^P, \tau)$ is an indicator function defined as:
    \begin{equation}
        I(w_i^G, w_j^P, \tau) =
    \begin{cases}
        1 & \text{if } cos(w_i^G, w_j^P) > \tau\\
        0 & \text{otherwise}
    \end{cases}
    \end{equation}
    We take $\tau = 0.7$ in our experiments.
    
    \item \textbf{Colours RGB distance}: Given the ground truth colour set $C^G = {c_1^G, c_2^G, ..., c_i^G}$ and predicted colour set $C^P = \{c_1^P, c_2^P, ..., c_j^P\}$, we map each colour to its RGB value to obtain the sets $W^G = \{w_1^G, w_2^G, ..., w_i^G\}$ and $W^P = \{w_1^P, w_2^P, ..., w_j^P\}$ where each element in the sets is a $3\times1$ dimensional vector of RGB values. For some distance threshold $\tau$, the mean euclidean distance is calculated as follows:
    \begin{equation}
        \frac{\mathop{\sum_{i=1}^{|C^G|}\sum_{j=1}^{|C^P|}} distance(w_i^G, w_j^P).\mathbb{I}(w_i^G, w_j^P, \tau)}{\mathop{\sum_{i=1}^{|C^G|}\sum_{j=1}^{|C^P|}} \mathbb{I}(w_i^G, w_j^P, \tau) }
    \end{equation}
    where $\mathbb{I}(w_i^G, w_j^P, \tau)$ is an indicator function defined as:
    \begin{equation}
        \mathbb{I}(w_i^G, w_j^P, \tau) =
    \begin{cases}
        1 & \text{if } distance(w_i^G, w_j^P) < \tau\\
        0 & \text{otherwise}
    \end{cases}
    \end{equation}
    We take $\tau = 0.5$ in our experiments.
    
    \item \textbf{Colours coverage RMSE}: Consider the intersection $I = C^G \cap C^P$ of ground truth and predicted colour sets. The root mean squared error between the area covered by colours present in both ground truth and predicted image is calculated as follows:
    \begin{equation}
        RMSE = \sqrt{\frac{1}{|I|}\sum_{i=1}^{|I|}(coverage(c_i^G) - coverage(c_i^P))^2}
    \end{equation}

    \item \textbf{Tones coverage RMSE}: Consider the intersection $I = T^G \cap T^P$ of ground truth and predicted image tones. The root mean squared error between the proportion of tones in ground truth and predicted image is calculated as follows:
    \begin{equation}
        RMSE = \sqrt{\frac{1}{|I|}\sum_{i=1}^{|I|}(coverage(t_i^G) - coverage(t_i^P))^2}
    \end{equation}

    \item \textbf{Objects IOU}: The intersection over union between set $O^G$ of objects in the ground truth image verbalization and set $O^P$ of objects in the predicted image verbalization is computed as:
    \begin{equation}
        IOU(O^G, O^P) = \frac{|O^G \cap O^P|}{|O^G \cup O^P|}
    \end{equation}
    
    \item \textbf{Objects similarity}: For the ground truth set of objects $O^G = \{o_1^G, o_2^G, ..., o_i^G\}$ and set of predicted objects $O^P = \{o_1^P, o_2^P, ..., o_j^P\}$, we correspondingly obtain the sets of word embeddings $W^G = \{w_1^G, w_2^G, ..., w_i^G\}$ and $W^P = \{w_1^P, w_2^P, ..., w_j^P\}$, using Spacy. For some similarity threshold $\tau$, the mean cosine similarity is computed as follows:
    \begin{equation}
        \frac{\mathop{\sum_{i=1}^{|O^G|}\sum_{j=1}^{|O^P|}} cos(w_i^G, w_j^P).\mathbb{I}(w_i^G, w_j^P, \tau)}{\mathop{\sum_{i=1}^{|O^G|}\sum_{j=1}^{|O^P|}} \mathbb{I}(w_i^G, w_j^P, \tau) }
    \end{equation}
    where $\mathbb{I}(w_i^G, w_j^P, \tau)$ is an indicator function defined as:
    \begin{equation}
        \mathbb{I}(w_i^G, w_j^P, \tau) =
    \begin{cases}
        1 & \text{if } cos(w_i^G, w_j^P) > \tau\\
        0 & \text{otherwise}
    \end{cases}
    \end{equation}
    We take $\tau = 0.7$ in our experiments.
    
    \item \textbf{Normalised objects area RMSE}: As described above, consider the sets of word vectors of objects present in the ground truth image $O^G = \{o_1^G, o_2^G, ..., o_i^G\}$ and predicted image $O^P = \{o_1^P, o_2^P, ..., o_j^P\}$. Given the ground truth image area $ A^G = width \times height$ and a similarity threshold $\tau$, we first compute the mean squared error between the areas of bounding boxes of similar objects in the ground truth and predicted image, weighted by the proportion of each object in the ground truth image and its cosine similarity with the object in the predicted image. Further, we take the square root of the error thus obtained and normalise it by $A^G$ to achieve the desired metric, as follows:

    \begin{equation}
        MSE = \frac{\mathop{\sum_{i=1}^{|O^G|}\sum_{j=1}^{|O^P|}} \{(area(o_i^G) - area(o_j^P))^2.\frac{area(o_i^G)}{A^G}.\frac{1}{cos(w_i^G, w_j^P)}\}.\mathbb{I}(w_i^G, w_j^P, \tau)}{\mathop{\sum_{i=1}^{|O^G|}\sum_{j=1}^{|O^P|}} \mathbb{I}(w_i^G, w_j^P, \tau) }
    \end{equation}
    \begin{equation}
        Normalised \text{ } RMSE = \frac{\sqrt{MSE}}{A^G}
    \end{equation}
    where $\mathbb{I}(w_i^G, w_j^P, \tau)$ is an indicator function as described above. We take $\tau = 0.7$ in our experiments.
    
    \item \textbf{Normalised relative position error}: Following a similar approach as explained above, we compute the mean euclidean distance between the centroids of bounding boxes of similar objects weighted by the cosine similarity of objects present in the ground truth and predicted images and normalise it by the length of diagonal in the ground truth image $D^G$:
    \begin{equation}
        RPE = \frac{\mathop{\sum_{i=1}^{|O^G|}\sum_{j=1}^{|O^P|}} \{(distance(centroid(o_i^G), centroid(o_j^P).\frac{1}{cos(w_i^G, w_j^P)}\}.\mathbb{I}(w_i^G, w_j^P, \tau)}{\mathop{\sum_{i=1}^{|O^G|}\sum_{j=1}^{|O^P|}} \mathbb{I}(w_i^G, w_j^P, \tau) }
    \end{equation}\
    \begin{equation}
        Normalised \text{ } RPE = \frac{RPE}{D^G}
    \end{equation}
    where $\mathbb{I}(w_i^G, w_j^P, \tau)$ is the aforementioned indicator function. As before, we take $\tau = 0.7$ in our experiments.
    
\end{itemize}

\subsection{Preparing the BOIG instructions dataset}
\label{sec:Preparing the BOIG instructions dataset}
\begin{lstlisting}[caption={Behavior Finetuning Verbalization Pattern (1): Explicitly asking model to pay attention to behaviour tokens},frame=single,label={BoigLLM:verbalization-1},basicstyle=\scriptsize]
Input: You are a smart model. I am giving you some data regarding an image - (1) captions (2) keywords (3) image resolution i.e. (width, height) (4) release date (5) number of downloads i.e. how many times the image was downloaded (6) number of forwards i.e. how many times the image was forwarded to someone else (7) number of impressions i.e. how many times the image was seen by someone. Note that (5), (6) and (7) are Key Performance Indicators (KPIs) of the image, thus they are important signals of its perceived quality and popularity.
You have to predict following attributes of the image: (1) colour and tones from the lists given below: - Allowed colours: ['Red', 'Dark_Red', 'Green', 'Bright_Green', 'Dark_Green', 'Light_Green', 'Mud_Green', 'Blue', 'Dark_Blue', 'Light_Blue', 'Royal_Blue', 'Black', 'White', 'Off_White', 'Gray', 'Dark_Gray', 'Silver', 'Cream', 'Magenta', 'Cyan', 'Yellow', 'Mustard', 'Khaki', 'Brown', 'Dark_Brown', 'Violet', 'Pink', 'Dark_Pink', 'Maroon', 'Tan', 'Purple', 'Lavender', 'Turquoise', 'Plum', 'Gold', 'Emerald', 'Orange', 'Beige', 'Lilac', 'Olive']  - Allowed tones: ['warm', 'neutral', 'cool'] (2) main objects present in the image and the diagonal coordinates of their bounding boxes: [x1, y1, x2, y2]
Now, predict the attributes for the following image: [captions: "Waist up portrait of mixed-race female worker posing confidently while standing with arms crossed in plant workshop", keywords: "female, worker, young, woman, mixed-race, african, african-american, modern, contemporary, work, occupation, industry, industrial, plant, factory, workshop, work shop, strong, tough, gritty, masculine, short, hair, latin-american, plump, adult, mechanic, repair, repairman, handywoman, foreman, copy space, portrait, looking at camera, standing, posing, smiling, recruitment, employment, job, opportunity, engineer, production, manufacturing, assembly, assembling, line", image resolution: "(5760, 3840)", release date: "2019-12-02", number of downloads: "24", number of forwards: "106", number of impressions: "5941"] Answer properly in JSON format. Do not include any other information in your answer.

Output: {"color and tones": {"colors": {"Gray": {"coverage": 0.4}, "Dark_Gray": {"coverage": 0.22}, "Black": {"coverage": 0.14}, "Off_White": {"coverage": 0.13}, "Silver": {"coverage": 0.11}}, "tones": {"warm": 0, "neutral": 1.0, "cool": 0}}, "objects": {"jeans": [2076.67, 2542.5, 3023.88, 3827.01], "woman": [1892.94, 11.18, 4260.09, 3824.34], "safety vest": [2160.75, 1410.95, 3668.16, 3826.63], "shirt": [2163.59, 1079.3, 4254.59, 3826.1]}}
\end{lstlisting}

\begin{lstlisting}[caption={Behavior Finetuning Verbalization Pattern (2): Noisy behavior in input and asking the model to correct the noise in addition to producing content},frame=single,label={BoigLLM:verbalization-2},basicstyle=\scriptsize]
Input: "You are a smart model. I am giving giving you some data regarding an image released by a content creator - (1) captions (2) keywords (3) image resolution i.e. (width, height) (4) release date (5) approximate number of downloads that the creator wants to achieve (6) approximate number of forwards that the creator wants to achieve (7) approximate number of impressions/views that the creator wants to achieve
You have to predict following attributes of the image: (1) colour and tones from the lists given below: - Allowed colours: ['Red', 'Dark_Red', 'Green', 'Bright_Green', 'Dark_Green', 'Light_Green', 'Mud_Green', 'Blue', 'Dark_Blue', 'Light_Blue', 'Royal_Blue', 'Black', 'White', 'Off_White', 'Gray', 'Dark_Gray', 'Silver', 'Cream', 'Magenta', 'Cyan', 'Yellow', 'Mustard', 'Khaki', 'Brown', 'Dark_Brown', 'Violet', 'Pink', 'Dark_Pink', 'Maroon', 'Tan', 'Purple', 'Lavender', 'Turquoise', 'Plum', 'Gold', 'Emerald', 'Orange', 'Beige', 'Lilac', 'Olive'] - Allowed tones: ['warm', 'neutral', 'cool']  (2) main objects present in the image and the diagonal coordinates of their bounding boxes: [x1, y1, x2, y2]  (3) exact number of downloads that the image will get  (4) exact number of forwards that the image will get  (5) exact number of impressions/views that the image will get
Now, predict the attributes for the following image: [ captions: ""Hispanic adult man holding 100 brazilian real banknotes smiling happy pointing with hand and finger to the side"", keywords: ""pointing, side, face, happy, hopeful, smile, finger, optimistic, hand, point, showing, looking, smiling, one, gesture, confident, up, cheerful, look, mouth, joy, friendly, expression, emotion, presentation, idea, blue, background, hispanic, latin, man, male, guy, beard, bald, shaved, adult, young, money, currency, business, brazilian, cash, brazil, real, investment, banknote, 100"", image resolution: ""(9216, 6144)"", release date: ""2021-02-27"", approximate number of downloads that the creator wants to achieve: ""4"", approximate number of forwards that the creator wants to achieve: ""17"", approximate number of impressions/views that the creator wants to achieve: ""919"" ] Answer properly in JSON format. Do not include any other information in your answer."

Output: {"color and tones": {"colors": {"Cyan": {"coverage": 0.69}, "Light_Blue": {"coverage": 0.16}, "Turquoise": {"coverage": 0.15}}, "tones": {"warm": 0, "neutral": 0, "cool": 1.0}}, "objects": {"man": [1068.6, 18.57, 8143.44, 6121.09], "banknote bill": [3443.5, 2146.79, 5294.15, 3455.05]}, "exact downloads": 4, "exact forwards": 15, "exact impressions": 885}

\end{lstlisting}

\begin{lstlisting}[caption={Behavior Finetuning Verbalization Pattern (3): Noisy behavior in input and asking the model to correct behavior in output},frame=single,label={BoigLLM:verbalization-3},basicstyle=\scriptsize]
Input: "You are a smart model. I am giving giving you some data regarding an image released by a content creator - (1) captions (2) keywords (3) image resolution i.e. (width, height) (4) release date (5) approximate number of downloads that the creator wants to achieve (6) approximate number of forwards that the creator wants to achieve (7) approximate number of impressions/views that the creator wants to achieve
You have to predict following attributes of the image: (1) exact number of downloads that the image will get (2) exact number of forwards that the image will get (3) exact number of impressions/views that the image will get.
Now, predict the attributes for the following image: [ captions: ""Movie slapstick vector illustration. Behind the scenes inscription on flapper"", keywords: ""behind the scenes, slapstick, flapper, movie, cinema, scene, logo, frame, film, duration, behind, act, black, cameraman, clip, date, director, entertainment, flap, footage, gray, hollywood, icon, illustration, inscription, operator, screen, shooting, sign, signal, symbol, television, theater, time, timecode, tv, vector, video, view, white"", image resolution: ""(4096, 4096)"", release date: ""2017-06-11"", approximate number of downloads that the creator wants to achieve: ""5"", approximate number of forwards that the creator wants to achieve: ""31"", approximate number of impressions/views that the creator wants to achieve: ""914"" ] Answer properly in JSON format. Do not include any other information in your answer. "

Output: {"exact downloads": 5, "exact forwards": 32, "exact impressions": 864}
\end{lstlisting}

\begin{lstlisting}[caption={Behavior Finetuning Verbalization Pattern (4): No behaviour in input and asking the model to simulate exact behaviour in output},frame=single,label={BoigLLM:verbalization-4},basicstyle=\scriptsize]
Input: "You are a smart model. I am giving giving you some data regarding an image released by a content creator - (1) captions (2) keywords (3) image resolution i.e. (width, height) (4) release date
You have to predict following attributes of the image: (1) exact number of downloads that the image will get (2) exact number of forwards that the image will get (3) exact number of impressions/views that the image will get
Now, predict the attributes for the following image: [captions: ""Company employees working in software development and designer office"", keywords: ""business, office, meeting, collaegue, successful, workplace, analysis, architect, coworker, discussion, entrepreneur, marketing, professional, company, employee, occupation, software, work, worker, team, people, brainstorming, cooperation, corporate, project, strategy, teamwork, together, computer, colleagues, young, diverse, collaboration, design, developer, group, ideas, management, smiling, multiethnic, place, plan, research, startup, technology, women, programmer, architects"", image resolution: ""(4035, 2690)"", release date: ""2020-09-29"" ]. Answer properly in JSON format. Do not include any other information in your answer."

Output: {"exact downloads": 1, "exact forwards": 1, "exact impressions": 186}
\end{lstlisting}

Hyperparameters for Behavior fine-tuning Llama: 
\begin{itemize}
    \item per device train batch size: 16
    \item gradient accumulation steps: 1
    \item max context length: 2048
    \item warmup ratio: 0.03
    \item warmup learning rate scheduler: cosine
    \item after warmup, learning rate: 2e-6
\end{itemize}

\begin{figure}[t]
    \centering
    \begin{subfigure}[b]{0.22\textwidth}
         \centering
         \includegraphics[width=1\textwidth,scale=0.9]{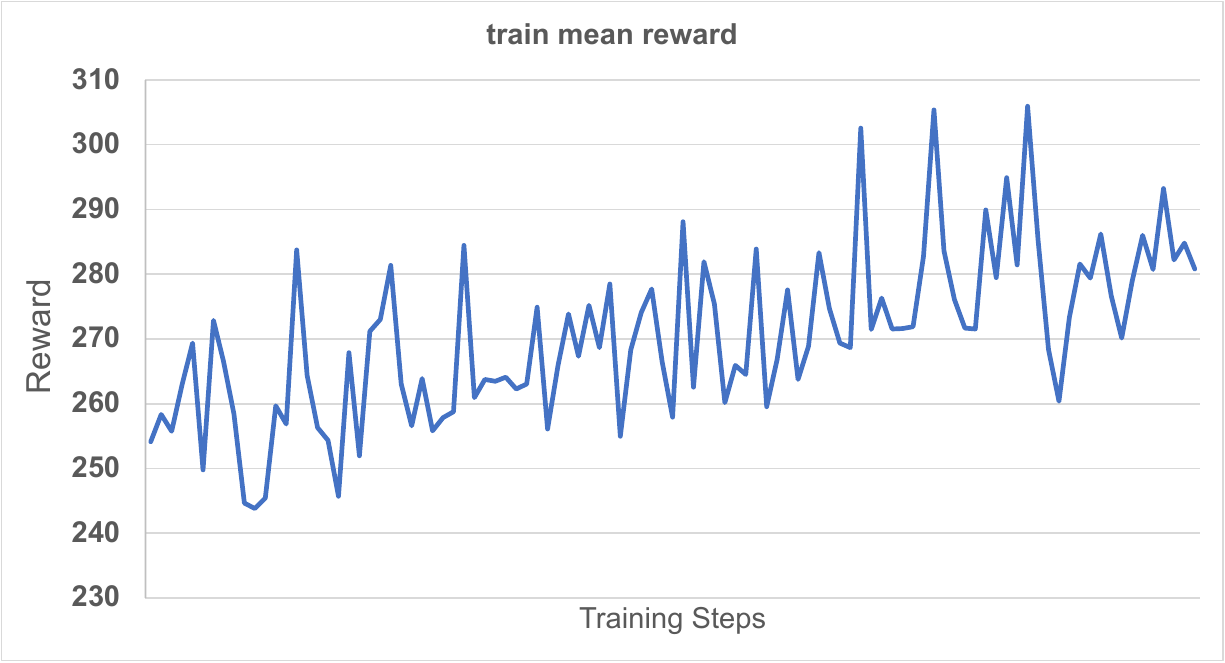}
         \caption{}
     \end{subfigure}
     \begin{subfigure}[b]{0.22\textwidth}
         \centering
         \includegraphics[width=1\textwidth,scale=0.9]{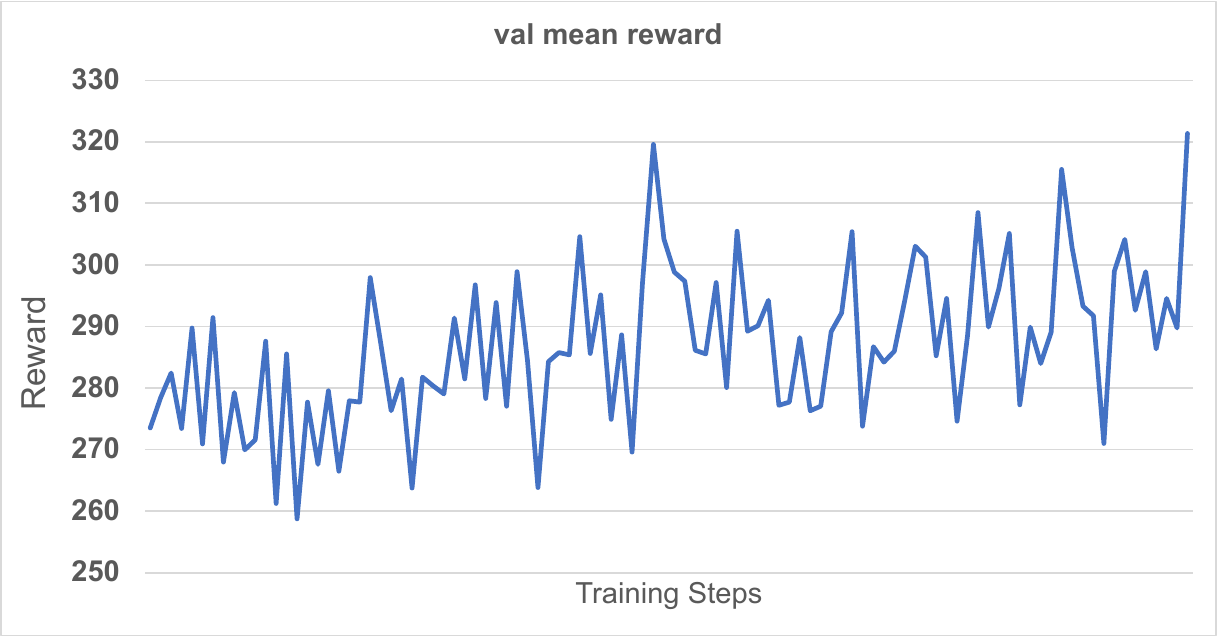}
         \caption{}
     \end{subfigure}    
    \begin{subfigure}[b]{0.22\textwidth}
         \centering
         \includegraphics[width=1\textwidth,scale=0.9]{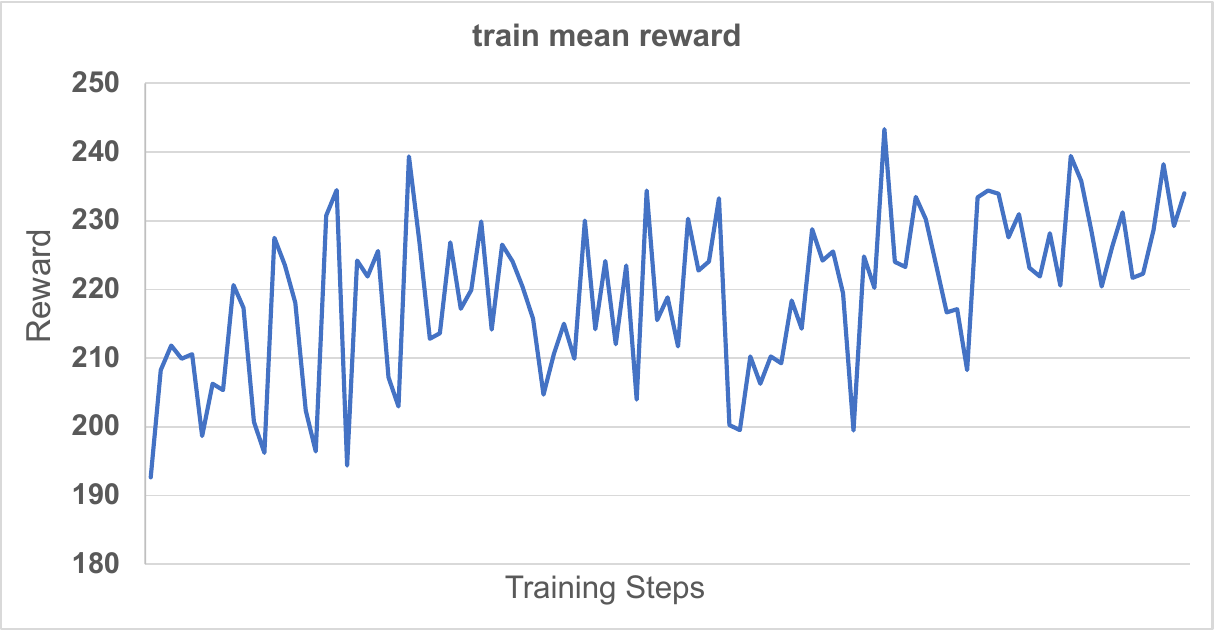}
         \caption{}
         \label{fig:stock_reward_curve_train}
     \end{subfigure}
     \begin{subfigure}[b]{0.22\textwidth}
         \centering
         \includegraphics[width=1\textwidth,scale=0.9]{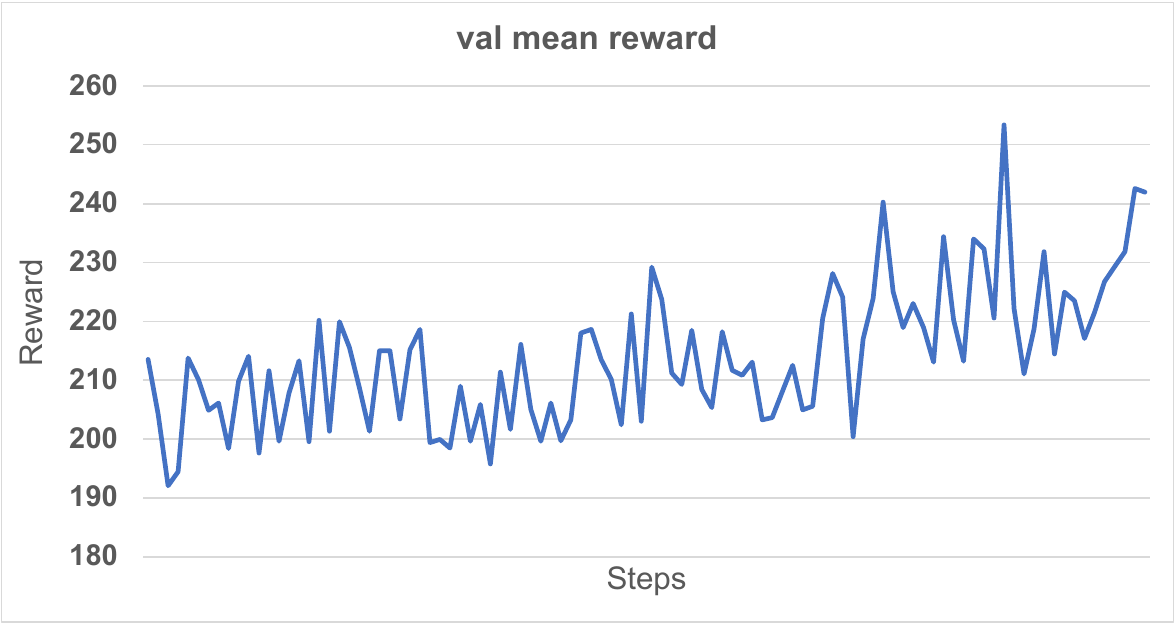}
         \caption{}
         \label{fig:stock_reward_curve_val}
     \end{subfigure}

     \caption{Reward curves for the performance alignment of stable diffusion on BoigBench (train (a) and validation (b) sets) and Stock datasets (train (c) and validation (d) sets)}
    \label{fig:stock_reward_curves}
\end{figure}

\section{BOIGBench Additional Details}
\label{sec:BOIGBench Additional Details}

\begin{table}[!htp]\centering
\caption{Distribution of ground truth BoigBench images }\label{tab:twitter_data_scores-ground-truth}
\scriptsize
\resizebox{0.4\textwidth}{!}{
\begin{tabular}{cccc}\toprule
\textbf{KPI}&\textbf{\# Objects} &\textbf{Aesthetic Score} &\textbf{CLIP Score} \\\midrule
High& 3.405 & 4.994 & 30.509\\
Low & 3.291 & 4.881 & 30.638\\
\bottomrule
\end{tabular}}
\end{table}

\begin{table}[!htp]\centering
\caption{Distribution of ground truth Stock images }\label{tab:stock_data_scores-ground-truth}
\scriptsize
\resizebox{0.4\textwidth}{!}{
\begin{tabular}{cccc}\toprule
\textbf{KPI} &\textbf{\# Objects} & \textbf{Aesthetic Score} &\textbf{CLIP Score} \\\midrule
High & 1.993 &5.339 &30.885\\
Medium & 2.043 & 5.351 &30.385\\
Low & 2.011 & 5.323 &30.523\\
\bottomrule
\end{tabular}}
\end{table}

\begin{figure}[t]
    \centering
    \begin{subfigure}[b]{0.22\textwidth}
         \centering
         \includegraphics[width=1\textwidth,scale=0.9]{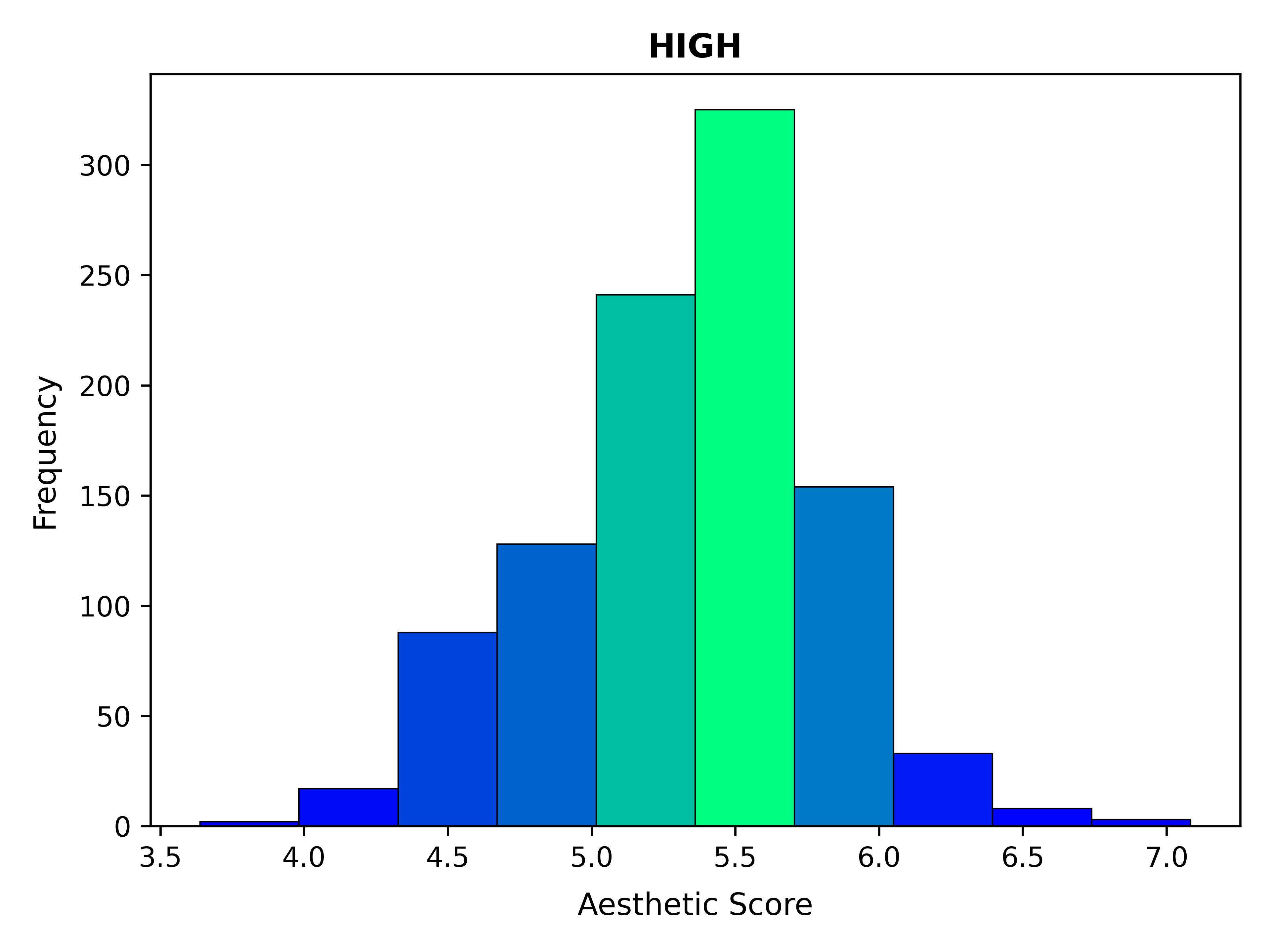}
         \caption{}
     \end{subfigure}
     \begin{subfigure}[b]{0.22\textwidth}
         \centering
         \includegraphics[width=1\textwidth,scale=0.9]{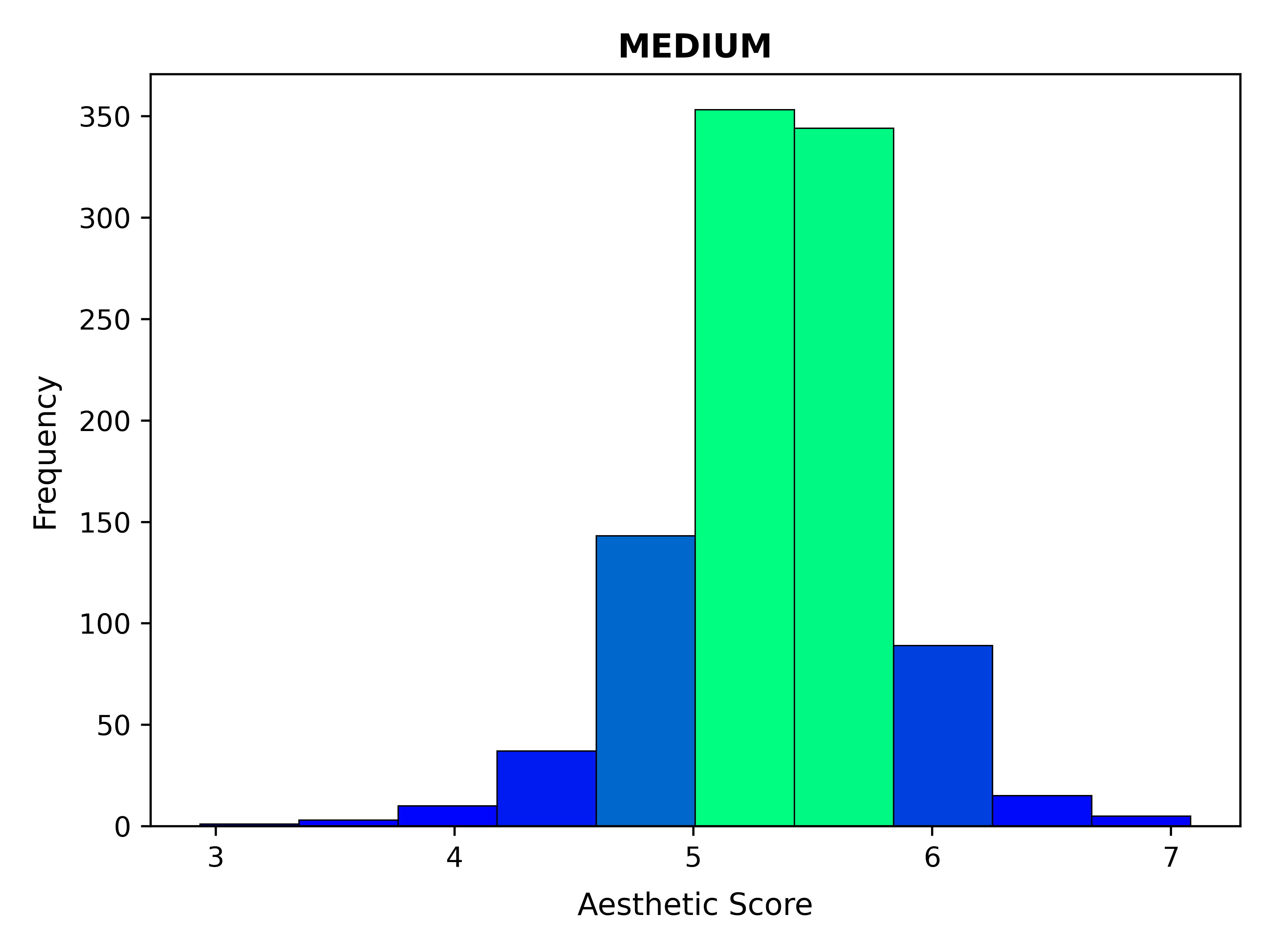}
         \caption{}
     \end{subfigure}
     \begin{subfigure}[b]{0.22\textwidth}
         \centering
         \includegraphics[width=1\textwidth,scale=0.9]{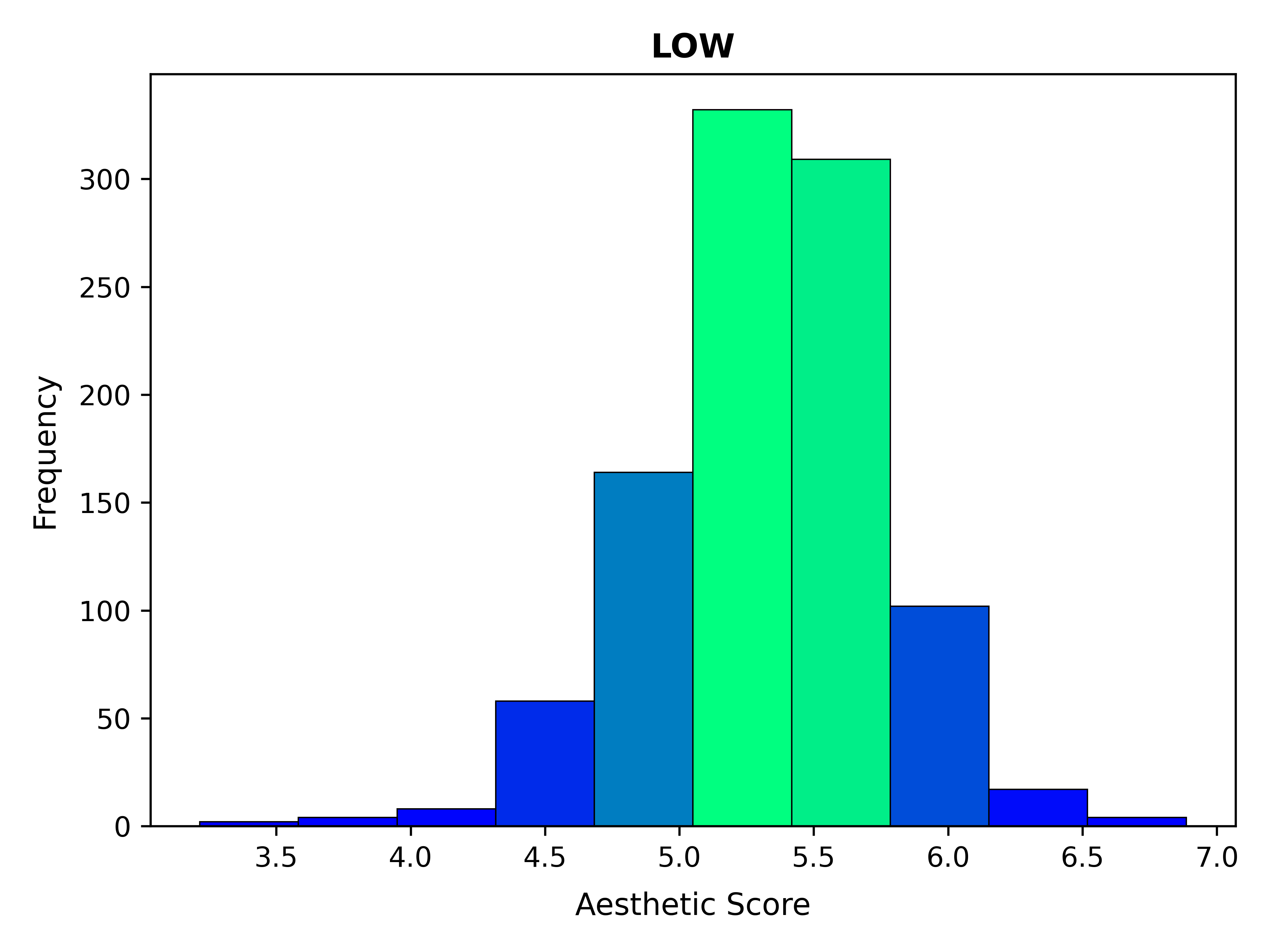}
         \caption{}
     \end{subfigure}

    \caption{Aesthetic Score distribution across High, Medium and Low KPI images in Stock dataset
    \label{fig:aesthetic_score_distribution_stock}}
\end{figure}

\begin{figure}[t]
    \centering
    \begin{subfigure}[b]{0.22\textwidth}
         \centering
         \includegraphics[width=1\textwidth,scale=0.9]{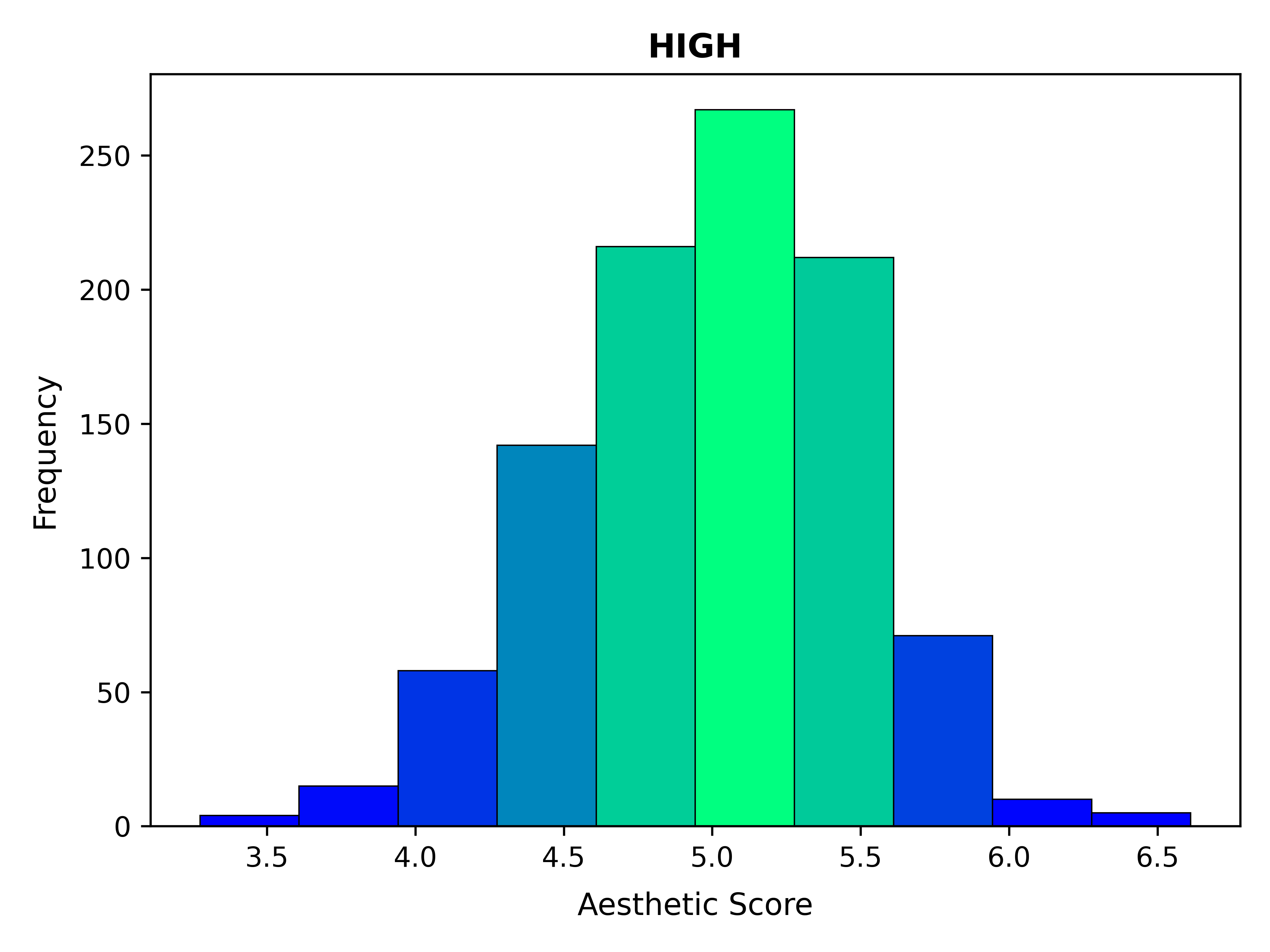}
         \caption{}
     \end{subfigure}
     \begin{subfigure}[b]{0.22\textwidth}
         \centering
         \includegraphics[width=1\textwidth,scale=0.9]{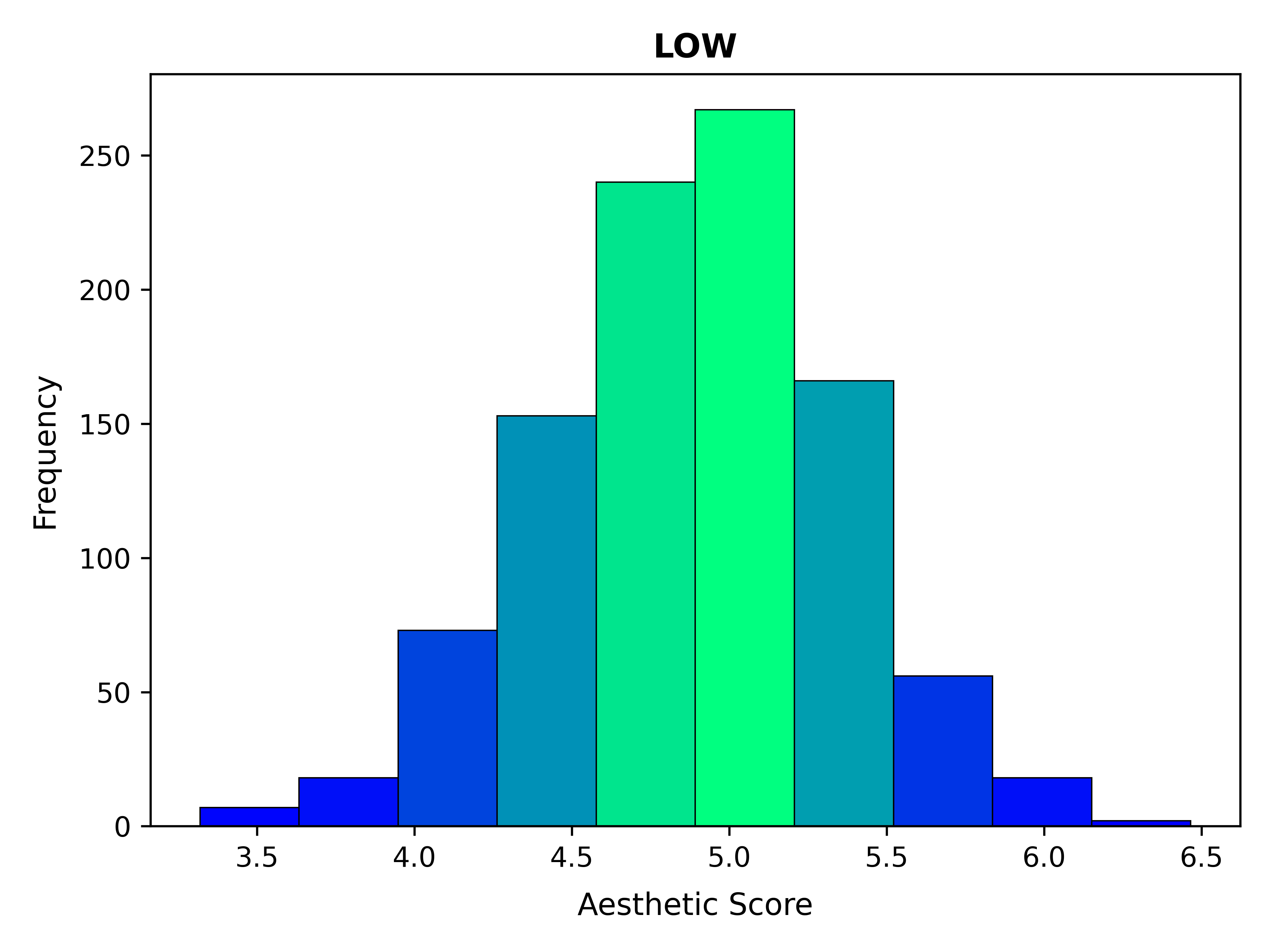}
         \caption{}
     \end{subfigure}

    \caption{Aesthetic Score distribution across High and Low KPI images in BoigBench dataset}
    \label{fig:aesthetic_score_distribution_boigbench}
\end{figure}

\begin{figure*}[t]
    \centering
    \begin{subfigure}[b]{0.45\textwidth}
         \centering
         \includegraphics[width=1\textwidth,scale=1]{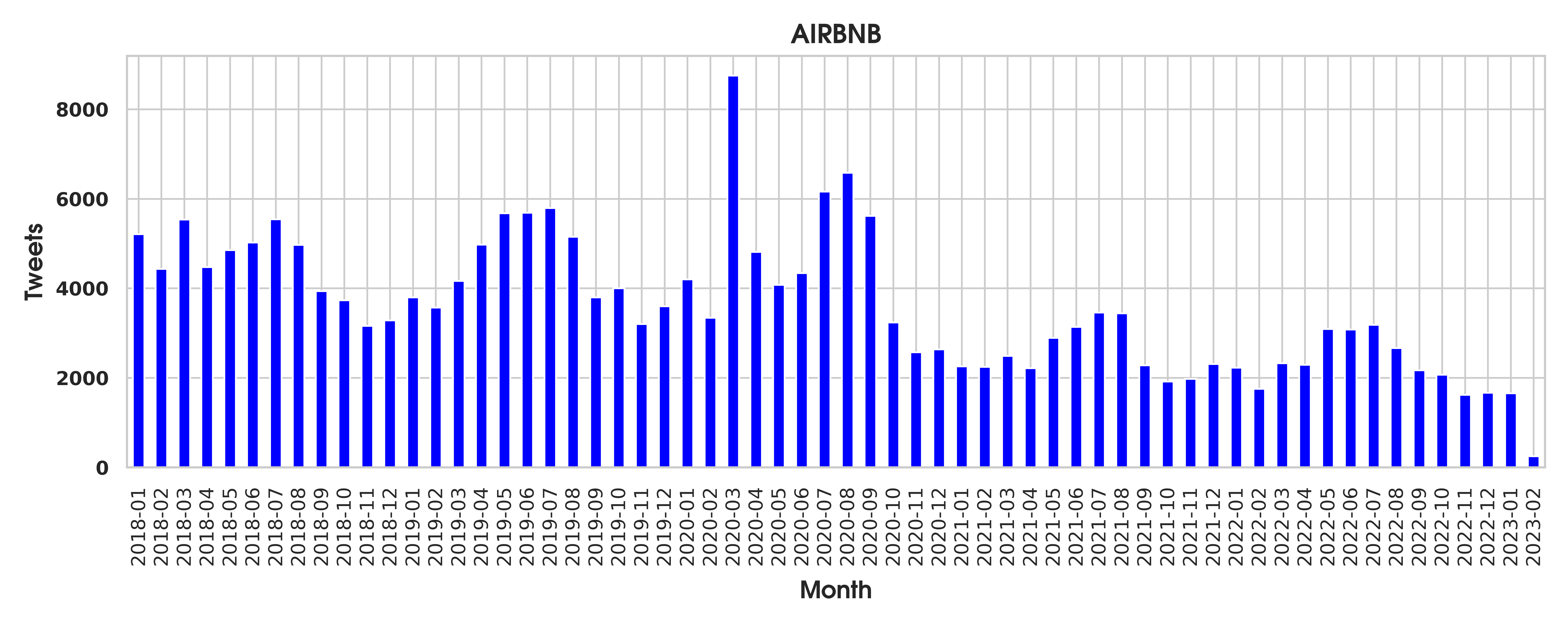}
         \caption{}
    \end{subfigure}
    \begin{subfigure}[b]{0.45\textwidth}
         \centering
         \includegraphics[width=1\textwidth,scale=1]{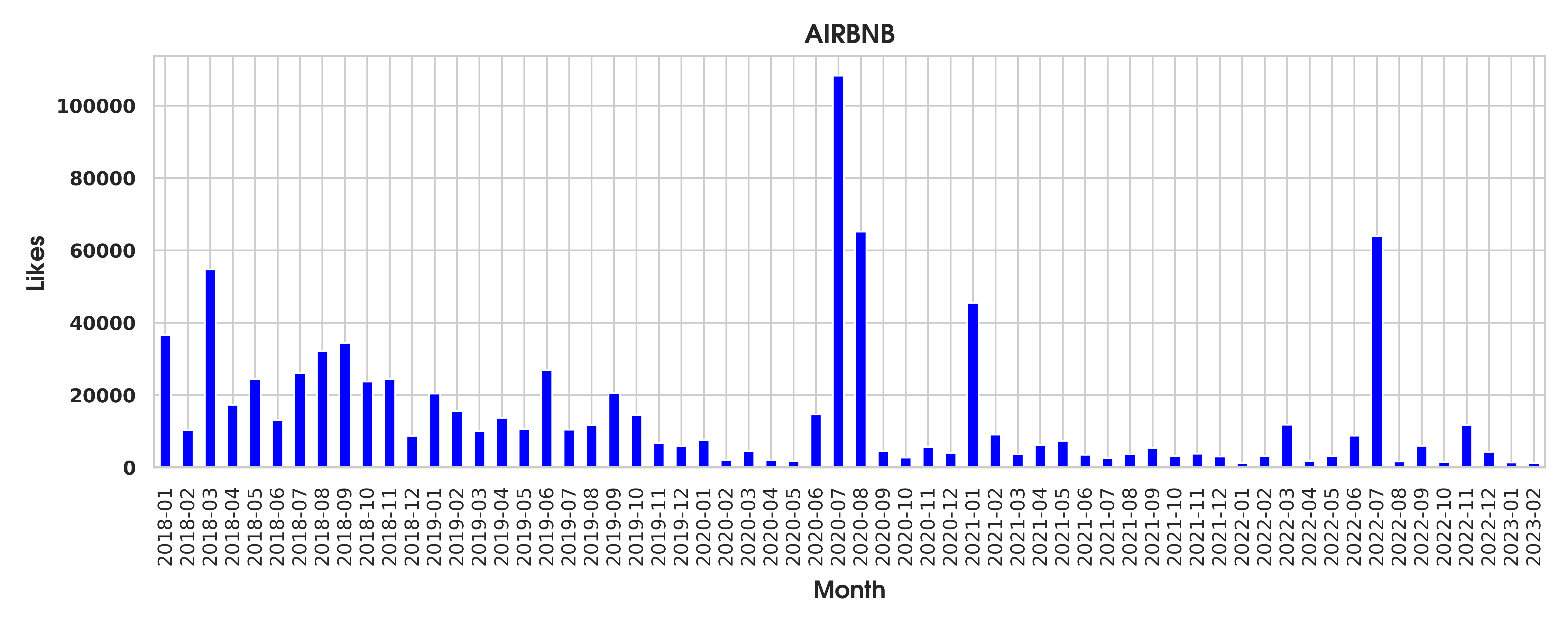}
         \caption{}
    \end{subfigure}
    \begin{subfigure}[b]{0.45\textwidth}
         \centering
         \includegraphics[width=1\textwidth,scale=1]{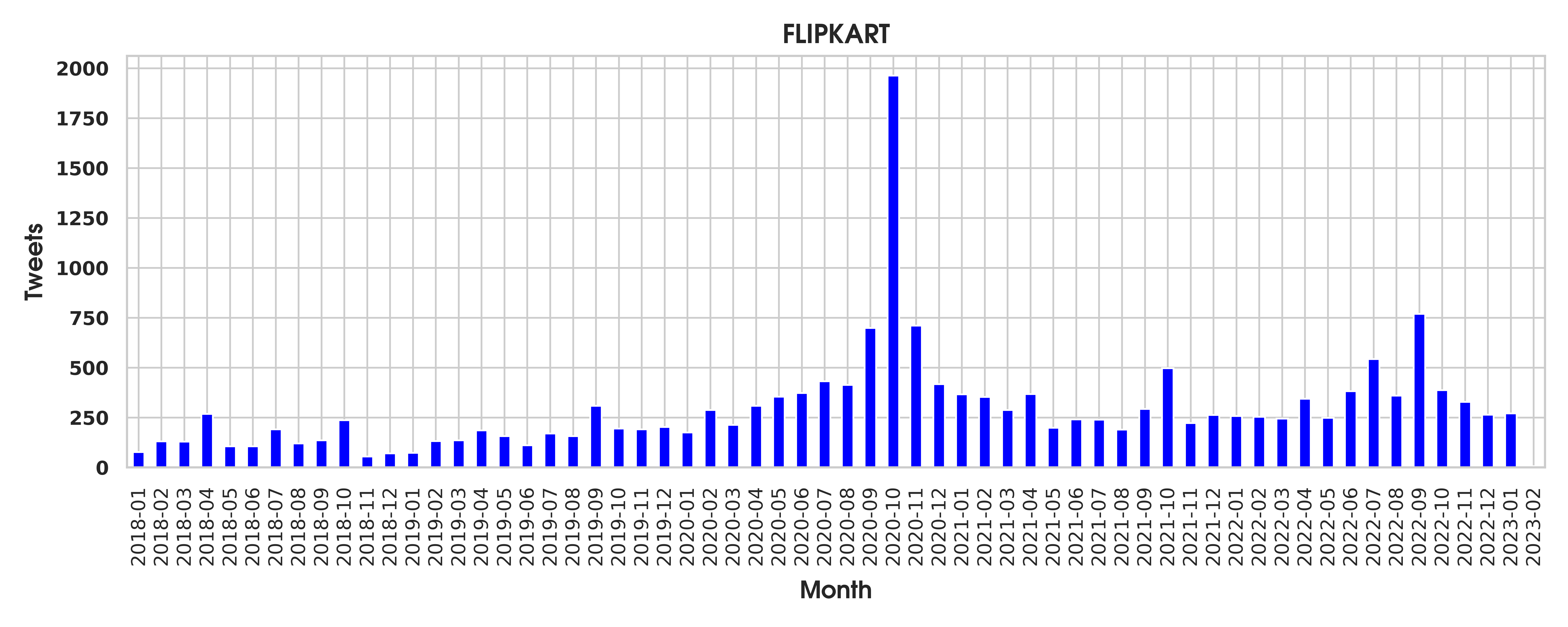}
         \caption{}
    \end{subfigure}
    \begin{subfigure}[b]{0.45\textwidth}
         \centering
         \includegraphics[width=1\columnwidth,scale=1]{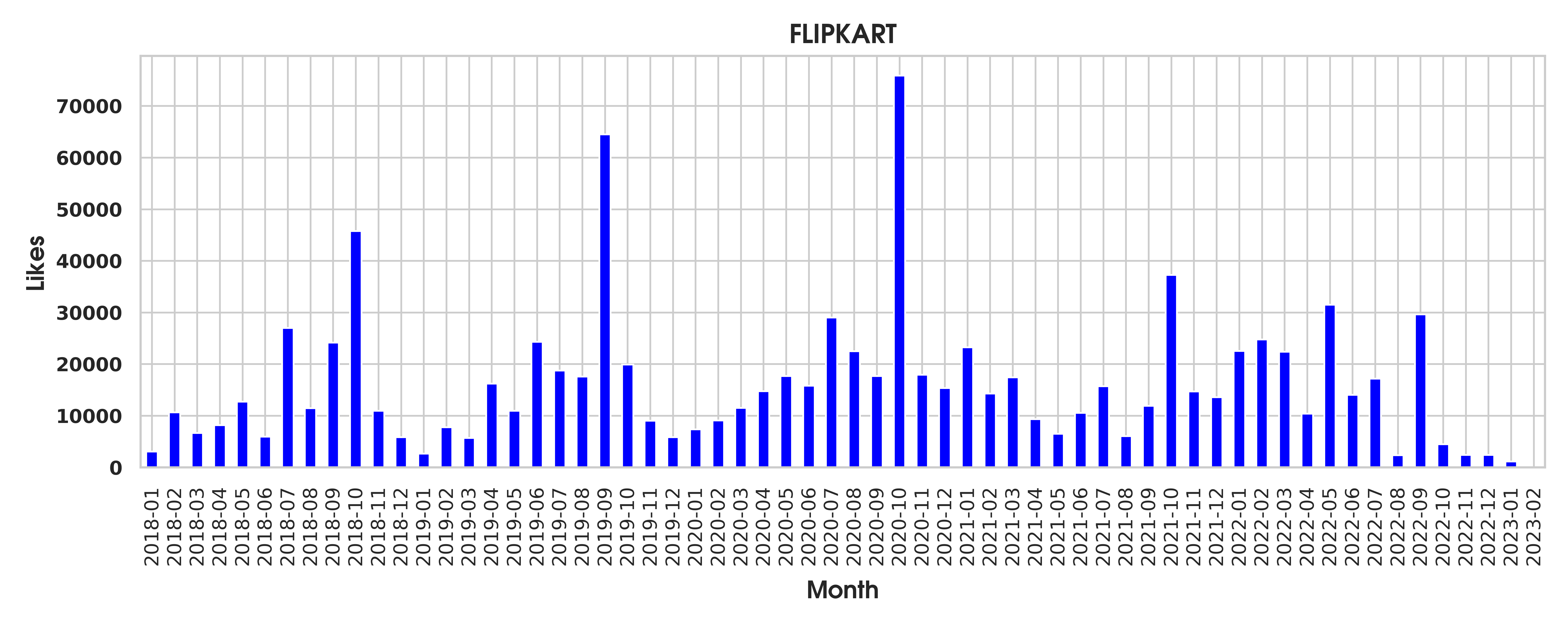}
         \caption{}
    \end{subfigure}
    \begin{subfigure}[b]{0.45\textwidth}
         \centering
         \includegraphics[width=1\columnwidth,scale=1]{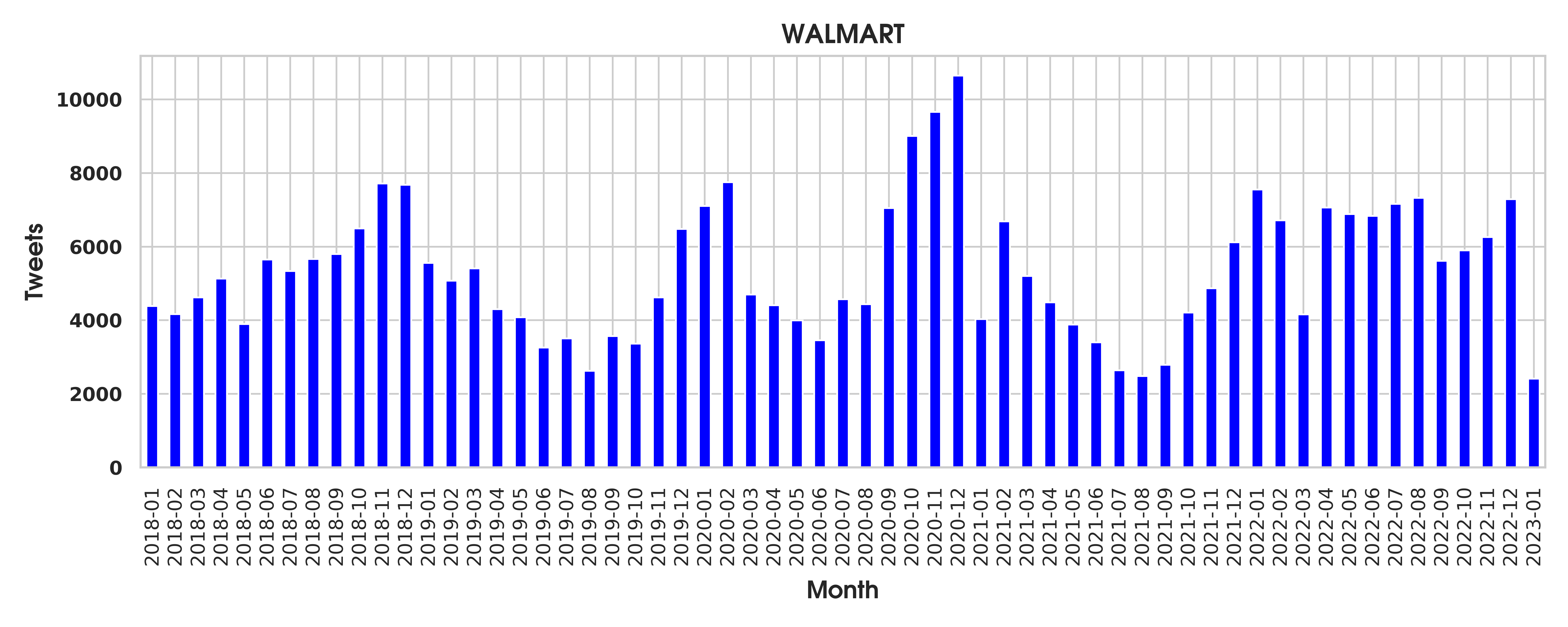}
         \caption{}
    \end{subfigure}
    \begin{subfigure}[b]{0.45\textwidth}
         \centering
         \includegraphics[width=1\columnwidth,scale=1]{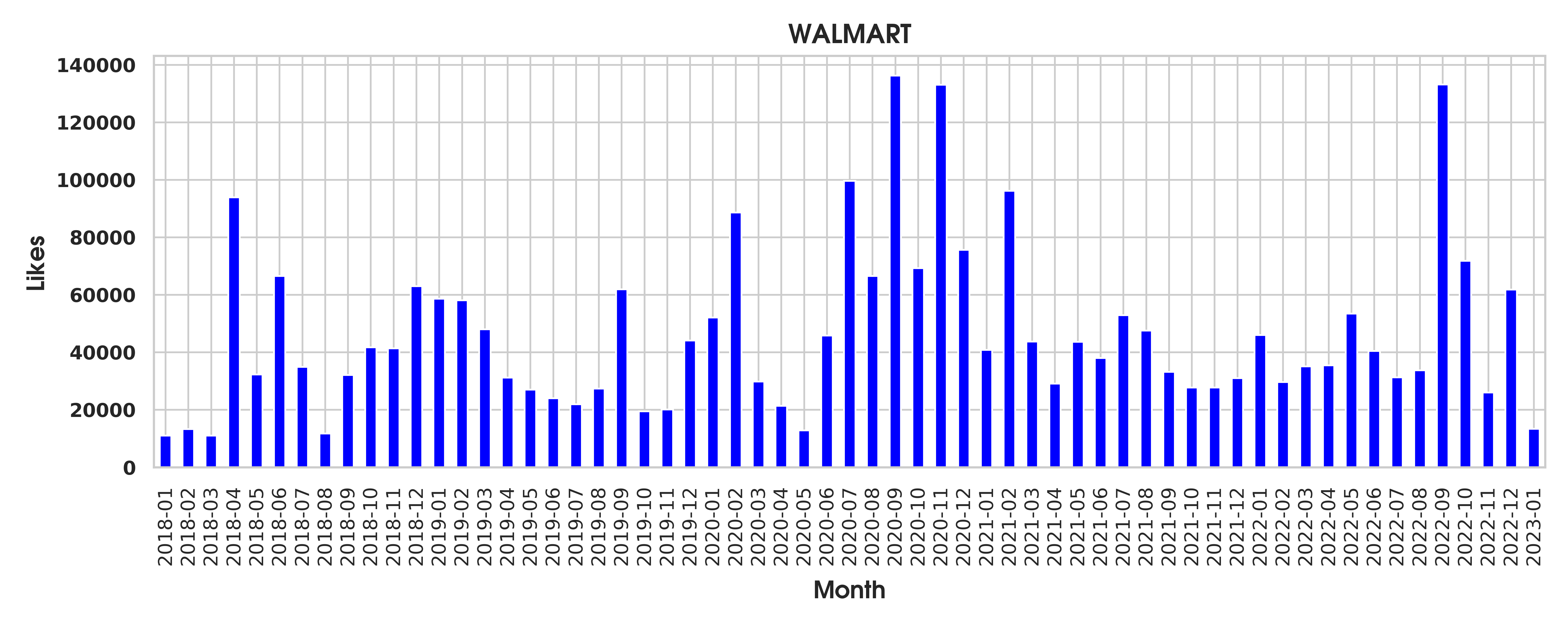}
         \caption{}
    \end{subfigure}
    \begin{subfigure}[b]{0.45\textwidth}
         \centering
         \includegraphics[width=1\columnwidth,scale=1]{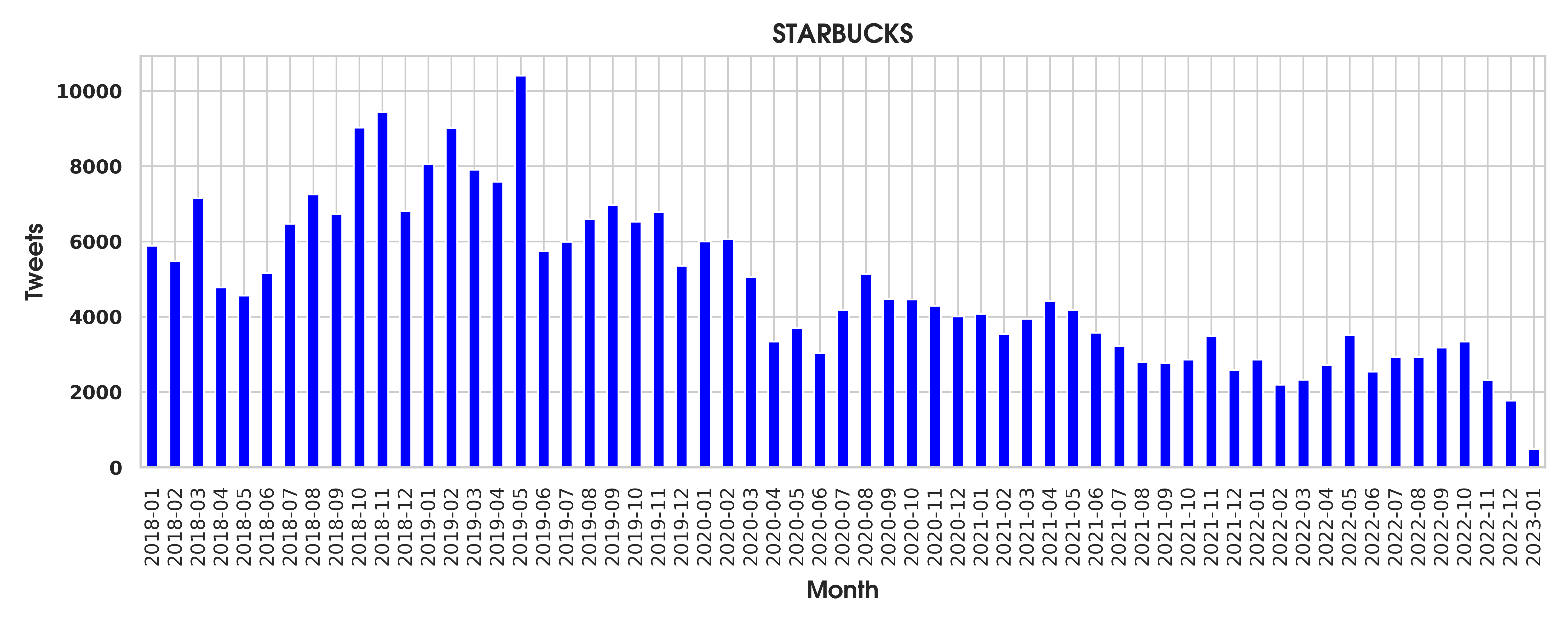}
         \caption{}
    \end{subfigure}
    \begin{subfigure}[b]{0.45\textwidth}
         \centering
         \includegraphics[width=1\columnwidth,scale=1]{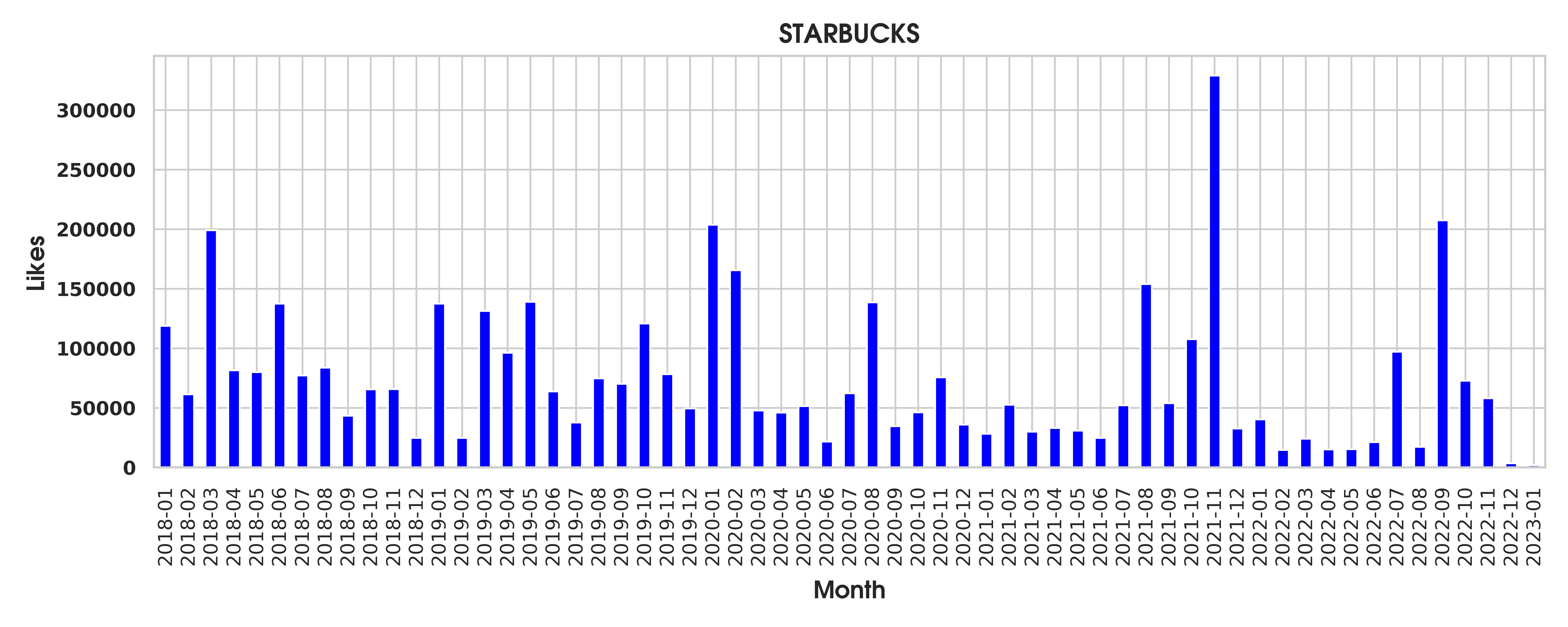}
         \caption{}
    \end{subfigure}
    \begin{subfigure}[b]{0.45\textwidth}
         \centering
         \includegraphics[width=1\columnwidth,scale=1]{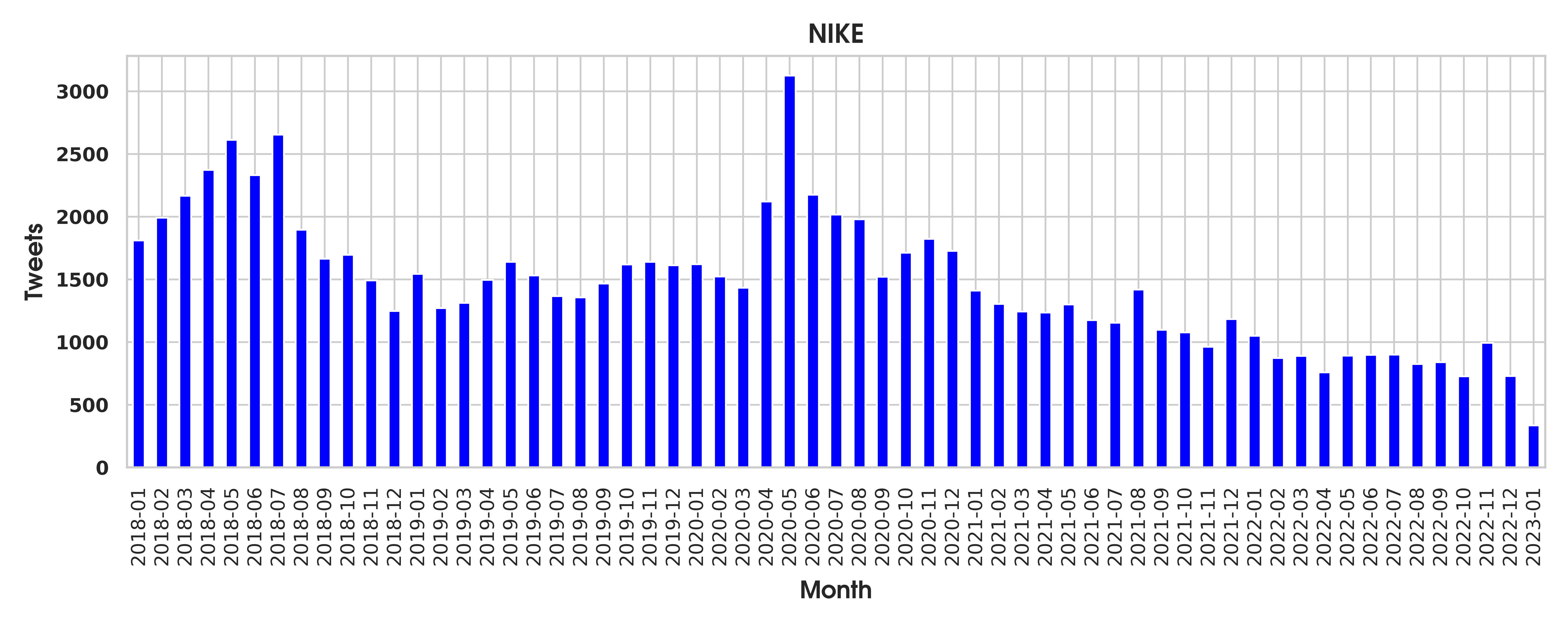}
         \caption{}
    \end{subfigure}
    \begin{subfigure}[b]{0.45\textwidth}
         \centering
         \includegraphics[width=1\columnwidth,scale=1]{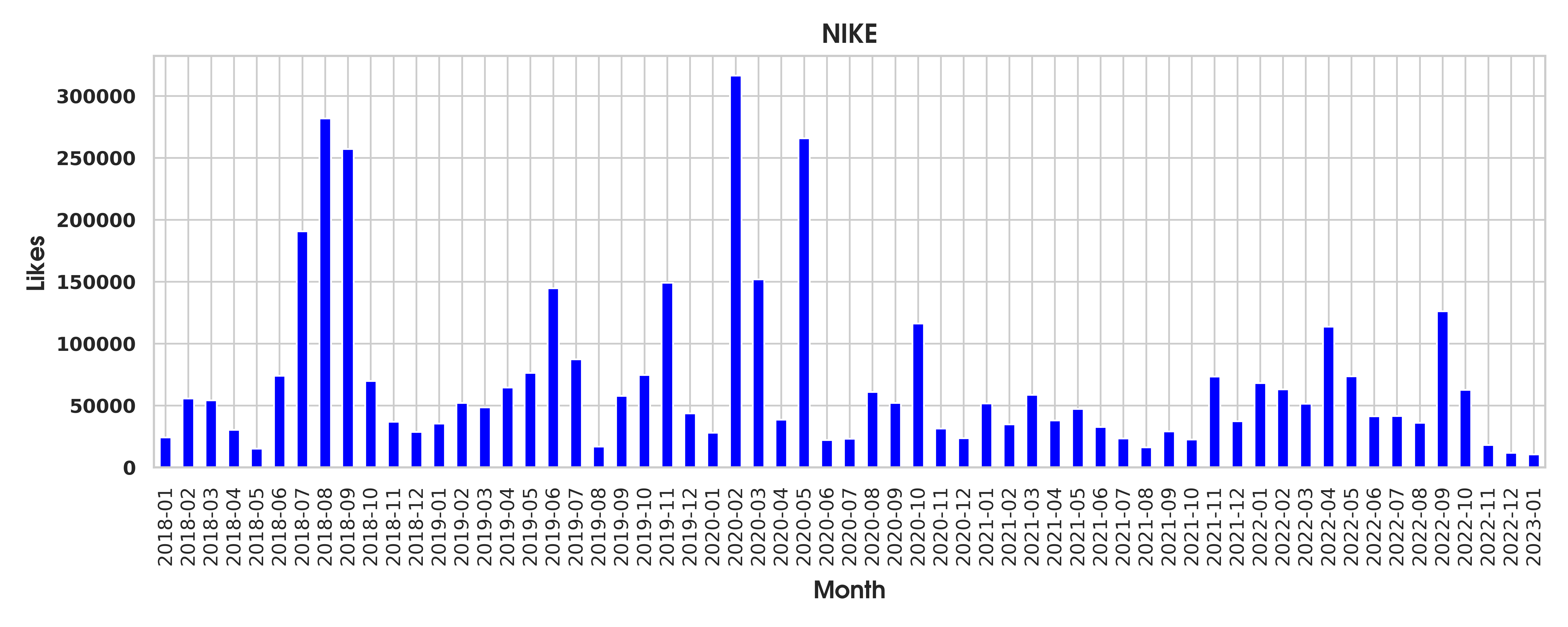}
         \caption{}
    \end{subfigure}
   
     \caption{Plots showing variation of number of tweets and likes with time for a few companies in the BoigBench dataset \label{fig:company_tweets_vs_time}}
\end{figure*}

\end{document}